\documentclass[journal,comsoc]{IEEEtran}
\usepackage[T1]{fontenc}
\usepackage{times}
\usepackage{epsfig}
\usepackage{graphicx}
\usepackage{amsmath}
\usepackage{amssymb}
\usepackage{bm}
\usepackage{algorithm}
\usepackage{algpseudocode}
\usepackage{color}

\DeclareMathOperator*{\argmin}{arg\,min}
\gdef\etal{et al.}

\usepackage{cite}

\ifCLASSINFOpdf
\else
\fi
\usepackage{amsmath}
\usepackage[cmintegrals]{newtxmath}
\begin{document}
%
\title{Robust Keyframe-based Dense SLAM with an RGB-D Camera}
%
%
%

\author{\begin{tabular}{cccccc}
	Haomin Liu$^{1\dagger}$, Chen Li$^{1\dagger}$, Guojun Chen$^{1}$, Guofeng Zhang$^{1*}$, Michael Kaess$^{2}$, and Hujun Bao$^{1}$
\end{tabular} \vspace{0.2cm}\\
\hspace{-0.3cm}$^1$State Key Lab of CAD\&CG, Zhejiang University
\hspace{0.5cm} $^2$Robotics Institute, Carnegie Mellon University
\thanks{$\dagger$ Joint first authors.}
\thanks{*~Corresponding author: Guofeng Zhang, Email: zhangguofeng@cad.zju.edu.cn}
}

%
%

\markboth{}%
{Shell \MakeLowercase{\textit{et al.}}: Bare Demo of IEEEtran.cls for IEEE Communications Society Journals}
%



\maketitle

\begin{abstract}
In this paper, we present RKD-SLAM, a robust keyframe-based dense SLAM approach for an RGB-D camera that can robustly handle fast motion and dense loop closure, and run without time limitation in a moderate size scene. It not only can be used to scan high-quality 3D models, but also can satisfy the demand of VR and AR applications. First, we combine color and depth information to construct a very fast keyframe-based tracking method on a CPU, which can work robustly in challenging cases (e.g.~fast camera motion and complex loops). For reducing accumulation error, we also introduce a very efficient incremental bundle adjustment (BA) algorithm, which can greatly save unnecessary computation and perform local and global BA in a unified optimization framework. An efficient keyframe-based depth representation and fusion method is proposed to generate and timely update the dense 3D surface with online correction according to the refined camera poses of keyframes through BA. The experimental results and comparisons on a variety of challenging datasets and TUM RGB-D benchmark demonstrate the effectiveness of the proposed system.
\end{abstract}


%
\IEEEpeerreviewmaketitle

\section{Introduction}
%
%
%
%
Simultaneous localization and mapping (SLAM) is a fundamental problem in both, the robotics and computer vision communities. Over the past decade, real-time structure-from-motion or visual SLAM has seen many successes~\cite{klein2007parallel,engel2014lsd,mur2015orb}. However, visual SLAM has inherent difficulty in handling textureless scenes and in reconstructing dense 3D information in real-time. Using depth sensors can help address these two problems. Along with the popularity of depth sensors~(e.g. Microsoft Kinect and Intel RealSense 3D Camera), more and more SLAM approaches~\cite{NewcombeIHMKDKSHF11,NiessnerZIS13,WhelanLSGD15,WhelanSGDL16,InfiniTAM_ISMAR_2015,WhelanKJFLM15} with depth or RGB-D sensors have been proposed.

Most dense SLAM methods use a frame-to-frame or frame-to-model alignment strategy, which easily results in accumulation of drift and fails eventually in challenging environments. Some methods~\cite{WhelanKJFLM15,WhelanLSGD15,WhelanSGDL16} proposed to use non-rigid mesh deformation techniques with loop closure constraints to optimize the map and limit drift. However, the model error caused by inaccurate alignment cannot be fully corrected by these methods, which may lead to increasing tracking error and eventual failure.

Recently, BundleFusion~\cite{dai2017bundle} was proposed, an end-to-end real-time 3D reconstruction system that uses all RGB-D input and globally optimizes the camera poses and 3D structure in an efficient hierarchical way. Different from previous methods using a frame-to-model strategy, BundleFusion performs brute-force matching for each input frame with all other frames, and then aligns the 3D points for fusion. However, it requires two powerful GPUs~(a NVIDIA GeForce GTX Titan X and a GTX Titan Black) to achieve real-time performance. Another major limitation is that it saves all RGB-D input data and only can run for about 10 minutes even with a very powerful PC, making it inappropriate for virtual reality and augmented reality applications which typically require much longer run time. Most recently, Maier~\etal~\cite{MaierSC17} proposed to improve this work by using a keyframe fusion and re-integration method based on DVO-SLAM~\cite{KerlSC13}, which can perform real-time dense SLAM with online surface correction using a single GPU.

In this paper, we present RKD-SLAM, a robust keyframe-based dense SLAM system for an RGB-D camera that is able to perform in real-time on a laptop without time limitation in a moderate size scene. RKD-SLAM also can handle fast camera motion and low-frame-rate live RGB-D sequences. The main contributions of our paper are as follows:

\begin{enumerate}
	\item We propose a robust keyframe-based RGB-D tracking method which combines visual and depth information to achieve robust and very fast camera tracking~(about $70\sim200$ fps) on a single CPU.
	
	\item We propose an efficient incremental bundle adjustment algorithm which makes maximum use of intermediate computation for efficiency, while adaptively updating affected keyframes for map refinement.
	
	\item We propose an efficient keyframe-based depth representation and fusion method which can generate and timely update the dense 3D surface with online correction without delay.
	
\end{enumerate}

\section{Related Work}
In the past few years, many methods have been proposed to use RGB-D camera data for dense 3D reconstruction and real-time SLAM. Huang~\etal~\cite{HuangBHKMFR11} proposed to use RGB-D data for real-time odometry, while dense mapping is done offline using sparse bundle adjustment~(BA). Endres~\etal~\cite{endres20143} presented a 3D mapping system using various visual features in combination with depth to estimate camera motion, while using 3D occupancy grid maps to represent the environment. Kerl~\etal~\cite{KerlSC13} proposed a dense direct RGB-D odometry by minimizing photometric error and depth error that leads to a higher pose accuracy when compared to sparse feature based methods. Newcombe~\etal~\cite{NewcombeIHMKDKSHF11} proposed an impressive dense SLAM system called KinectFusion, which used an iterative closest point~(ICP) algorithm~\cite{BeslM92} to align each frame to the global model with volumetric fusion. KinectFusion works well in a small scene, but could not handle a large-scale scene because in large-scale scenes the memory requirements for the volumetric space representation quickly exceeds any available memory. In addition, it suffers from drift problems and cannot handle loop closure. Following KinectFusion, many methods have been proposed to address these two problems. Most of them~\cite{Whelan12rssw,ZengZZL13,NiessnerZIS13} focused on exploiting more effective data structures for real-time volumetric fusion in a larger scale scene. For example, Kintinuous~\cite{Whelan12rssw} extends the KinectFusion with volume shift. Nie${\ss}$ner~\etal~\cite{NiessnerZIS13} proposed to use a sparse volumetric grid to store the volumetric information with spatial hashing. However, the drift problem of pose estimation and online dense surface adjustment are not addressed in these methods.

Drift-free pose estimation and sparse mapping have been extensively studied in visual SLAM. Many monocular SLAM methods have been proposed which can perform real-time tracking and sparse mapping in a small workspace~\cite{klein2007parallel} or even a street-scale scene~\cite{engel2014lsd,mur2015orb}. Relocalization and loop closure also can be handled online by some methods~\cite{engel2014lsd,mur2015orb,LiuZB16}. However, these methods do not generate dense 3D models. Although some methods~\cite{NewcombeLD11,PradeepRIZBB13,SchopsSHP15,OndruskaKI15} have been proposed to reconstruct dense 3D models online, it is still either limited to a small scene, or drift-free dense 3D reconstruction is not considered.

Although some offline methods~\cite{ZhouMK13,WangZB14,ChoiZK15} can close loops to obtain a drift-free dense 3D reconstruction, the computation speed is still far from real-time. Recently, Whelan~\etal\cite{WhelanKJFLM15} proposed a novel real-time dense RGB-D SLAM system with volumetric fusion, which can detect loop closure in a large-scale scene and correct drift through as-rigid-as-possible surface deformation. Instead of volumetric fusion, ElasticFusion~\cite{WhelanSGDL16} employed a surfel-based fusion method and also used the non-rigid surface deformation technique for loop closure and model refinement. Both of these methods use frame-to-model alignment, where the alignment error will affect the model accuracy, and the erroneous model will significantly harm the camera tracking. Using surface deformation with loop closure constraints cannot correct this error, so the tracking will eventually fail in complex environments. BundleFusion~\cite{dai2017bundle} uses brute-force matching to register frames, and re-integrates the depth maps of adjusted frames to obtain a globally consistent reconstruction. However, BundleFusion saves all input data, which is intractable for processing long sequences. In addition, the computation of brute-force matching also will become too time consuming for processing a very long sequence even with very powerful GPUs. Most recently, Maier~\etal~\cite{MaierSC17} proposed to use a keyframe fusion and re-integration strategy which can efficiently perform surface correction on-the-fly. In our system, we adopt this strategy with further improvement to perform depth fusion more timely. 

Bundle Adjustment~\cite{Triggs99} or pose graph optimization~\cite{OlsonLT06,StrasdatMD10,KummerleGSKB11} is frequently used in SLAM or SfM systems to reduce accumulation error or close loops to eliminate reconstruction drift.
Several works exploited the incremental nature of SLAM to speed up BA or smoothing~\cite{kaess2008isam,kaess2012isam2,ila2017slam++,IlaPSI17}. Instead of constructing and factorizing the information matrix from scratch for each incoming frame, Kaess~\etal~\cite{kaess2008isam} proposed to incrementally update the QR factorization of the information matrix. Later, Kaess~\etal~\cite{kaess2012isam2} further improved this method by using the Bayes tree to efficiently identify the subset of variables and the part of factorization that need to be updated. In \cite{ila2017slam++}, a similar method was proposed to update the mean of variables. In addition, it proposes an efficient method to update the covariance of variables. Most recently, Ila~\etal~\cite{IlaPSI17} proposed to incrementally update Schur complement for achieving fast incremental bundle adjustment. The covariance matrix is also efficiently recovered in their method.

\section{Framework Overview}
\begin{figure}
	\includegraphics[width=1.0\linewidth]{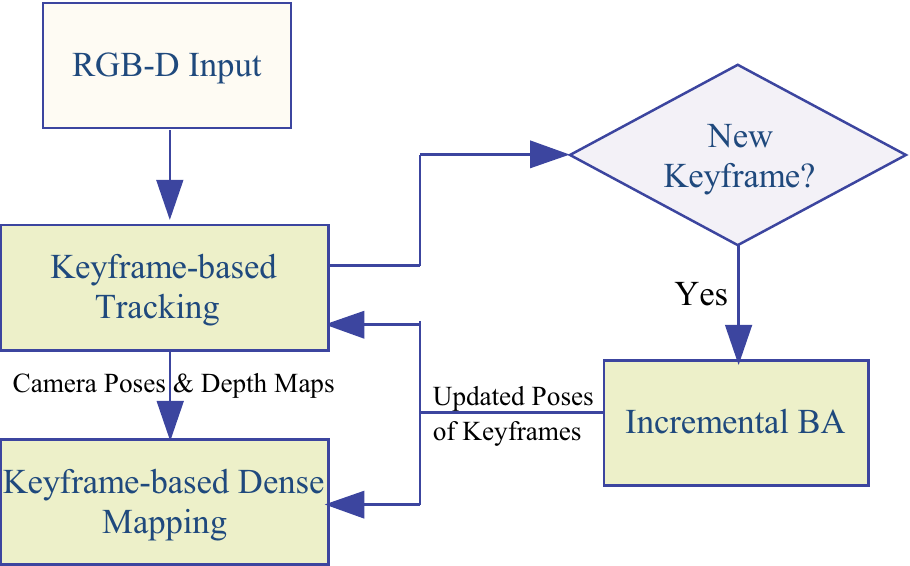}
	\caption{Framework of our system}
	\label{fig:framework}
\end{figure}

\begin{figure*}
	\includegraphics[width=1.0\linewidth]{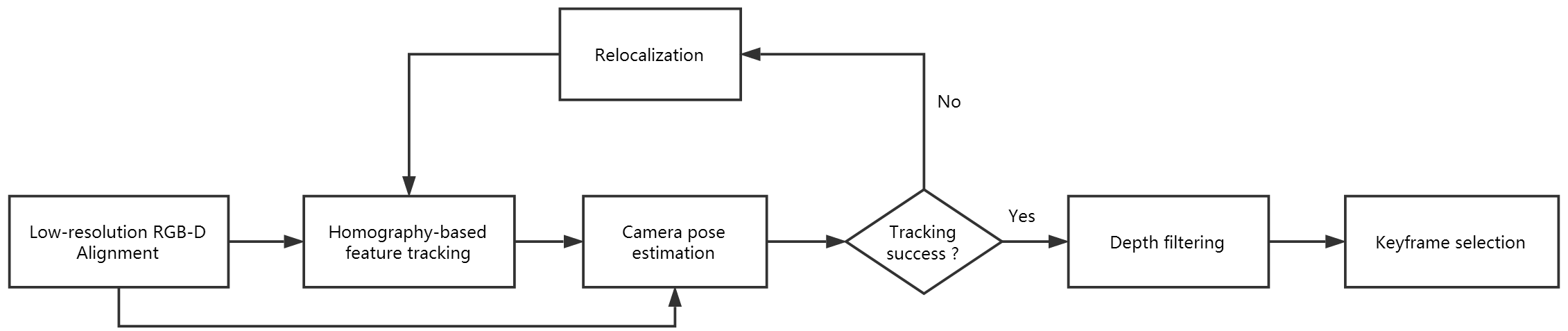}
	\caption{Framework of our keyframe-based tracking.}
	\label{fig:framework-odometry}
\end{figure*}

Figure~\ref{fig:framework} illustrates the framework of our system, which performs tracking and mapping in parallel threads. For each input frame, which contains an RGB image and a depth image, our system combines RGB-D alignment and a homography-based feature tracking method with depth filtering to make camera tracking as robust as possible. We also extract ORB features~\cite{rublee2011orb} for keyframes and match them with bags-of-words place recognition method~\cite{Galvez-LopezT12} to detect loop closure and build loop constraints. Periodically, an incremental BA is invoked in the background to refine the camera poses of keyframes and sparse 3D points. The mapping component will fuse the depth map of each input frame if its camera pose is accurately estimated. In order to allow adjusting of the dense surface online, the depth map is first fused to the keyframe with largest overlapping, which is followed by de-integration and integration in keyframes. This strategy allows our system to only re-integrate the depth maps and associated 3D point cloud of keyframes whose camera poses are refined by BA. By controlling the number of keyframes in a moderate size scene, our system can run in real-time without time limitation, even on a laptop.

\section{Keyframe-based Tracking}

Our keyframe-based tracking leverages both intensity and depth information to track camera motion for each frame. We combine both dense RGB-D based and sparse feature based methods to make the odometry more robust. The framework is illustrated in Figure~\ref{fig:framework-odometry}. 

\subsection{Feature Tracking with Low-resolution RGB-D Alignment}
For each current frame $F_i$~(its camera pose is denoted as ${\bf C}_i$), we first use a fast RGB-D alignment algorithm to estimate the relative pose ${\bf T}_{(i-1)\rightarrow i}$ from last frame $F_{i-1}$. Here we only need a coarse estimate of ${\bf T}_{(i-1)\rightarrow i}$, so we use ``small blurry image'' (SBI) as used in~\cite{klein2008improving} to achieve strong real-time performance without GPU acceleration. 

Similar to \cite{KerlSC13}, we project the previous frame $F_{i-1}$ to current frame $F_i$ and estimate the relative pose ${\bf T}_{(i-1)\rightarrow i}$ by solving the following energy function combining photometric error and inverse depth error
\begin{equation}
\small
\begin{aligned}
{\bf T}_{(i-1)\rightarrow i} = \argmin_{\bf T} \sum_{{\bf x} \in \tilde I_{i-1}} (||\frac{\tilde I_i({\bm\pi}({\bf K}{\bf T}(\tilde Z_{i-1}({\bf x}){\bf K}^{-1}\hat{\bf x}))) - \tilde I_{i-1}({\bf x})}{\sigma_c}||_{\delta} + &\\
||\frac{\tilde Z^{-1}_{i}({\bm\pi}({\bf K}{\bf T}(\tilde Z_{i-1}({\bf x}){\bf K}^{-1}\hat{\bf x}))) - z^{-1}({\bf T}(\tilde Z_{i-1}({\bf x}){\bf K}^{-1}\hat{\bf x}))}{\sigma_z}||_{\delta}),
\end{aligned}
\label{eq:RGBD-alignment}
\end{equation}
\normalsize
where $\tilde I_{i-1}$ and $\tilde I_i$ are the SBIs of $F_{i-1}$ and $F_{i}$ respectively. $\tilde Z_{i-1}$ and $\tilde Z_i$ are the downsampled depth maps of $F_{i-1}$ and $F_{i}$ respectively. $\hat{\bf x}$ denotes the homogenous coordinate of ${\bf x}$, and $z({\bf X})$ extracts the $z$-component of ${\bf X}$.
${\bm\pi}(\cdot)$ is the projective function ${\bm\pi}([x,y,z]^\top)=[x/z,y/z]^\top$. $\sigma_c$ and $\sigma_z$ are the parameters controlling corresponding weights. $||{\bm\epsilon}||_\delta$ is the Huber norm
\begin{equation}
||{\bm\epsilon}||_\delta = \left\{
\begin{array}{lll}
||{\bm\epsilon}||^2&&\text{if }||{\bm\epsilon}|| \leq \delta,\\
\delta(2||{\bm\epsilon}|| - \delta)&&\text{otherwise}.
\end{array} \right.\nonumber
\end{equation}
In our experiments, we generally set the Huber threshold $\delta = 1.345$.

An accurate pose estimate is obtained by feature correspondences. We track the map features in keyframes to the current frame by the homography-based feature tracking proposed in \cite{LiuZB16} to handle strong rotation and fast motion. For reducing computation, we only use global homography for tracking. During strong rotation and fast motion, the perspective distortion between frames may be too large for robust feature matching. Homography helps to rectify the patch distortion, so that the simple zero-mean SSD~\cite{klein2007parallel} is able to work.

As in \cite{LiuZB16}, we propagate the global homography from the last frame to the current frame. For a keyframe $k$, the global homography is propagated as ${\bf H}_{k\rightarrow i} = {\bf H}_{(i-1)\rightarrow i} {\bf H}_{k\rightarrow (i-1)}$, where ${\bf H}_{(i-1)\rightarrow i}$ is obtained by direct image alignment using the small blurry image (SBI) as used in~\cite{klein2008improving}:
\begin{equation}
{\bf H}_{(i-1)\rightarrow i}=\argmin_{{\bf H}} \sum_{{\bf x} \in {\tilde I_{i-1}}} ||\frac{\tilde I_{i-1}({\bf x}) - \tilde I_i(\tilde{\bf H}{\bf x})}{\sigma_c}||_{\delta},
\end{equation}
where $\tilde I_{i-1}$ and $\tilde I_i$ are the SBIs of last frame and the current frame respectively. The tilde above the homography $\tilde{\bf H}$ converts ${\bf H}$ from the original image space to that of SBI. After obtaining a set of feature matches $\mathcal{M}_{k\rightarrow i} = \{({\bf x}_k, {\bf x}_i)\}$ between keyframe $k$ and current frame $i$, we refine ${\bf H}_{k\rightarrow i}$ by:
\begin{equation}
\begin{aligned}
{\bf H}_{k\rightarrow i} = \argmin_{{\bf H}}\sum_{{\bf x} \in {\tilde I_k}} ||\frac{\tilde I_k({\bf x}) - \tilde I_i(\tilde{\bf H}{\bf x})}{\sigma_c}||_{\delta}&\\
+\sum_{({\bf x}_k,{\bf x}_i)\in \mathcal{M}_{k\rightarrow i}}\frac{1}{\sigma^2_{\bf x}}||{\bf H}{\bf x}_k - {\bf x}_i||_2^2&.
\end{aligned}
\end{equation}
Incorporating $\mathcal{M}_{k\rightarrow i}$ into direct image alignment can prevent the solution being biased towards a major plane.

Note that we only need to track a small set of keyframes to the current frame. Similar to \cite{mur2015orb}, we first select the set $\mathcal{K}_1$ containing keyframes sharing common points with the last frame, then select the second set $\mathcal{K}_2$ sharing common points with keyframes in $\mathcal{K}_1$.

Homography is also used to determine a small search region around ${\bf x}_i^\text{homo}$ instead of searching along the whole epipolar line. In this work, since we have depth measurement $z_k$ and the estimated camera pose obtained by RGB-D alignment ${\bf C}_i = {\bf T}_{(i-1)\rightarrow i} {\bf C}_{i-1}$, we define the search region as the union of the one around ${\bf x}_i^\text{homo}$ and the one around ${\bf x}_i^\text{RGB-D} = {\bm\pi}({\bf K}({\bf C}_i {\bf C}_k^{-1} (z_k {\bf K}^{-1}\hat{\bf x}_k)))$. Here $\hat{{\bf x}}_k$ denotes the homogenous coordinate of ${\bf x}_k$, and ${\bf K}$ is the intrinsic matrix which is assumed to be known and constant.

With the relative pose ${\bf T}_{(i-1)\rightarrow i}$ estimated by low-resolution RGB-D alignment and the set of 3D-2D feature correspondences $\mathcal{X} = \{({\bf X}_j, {\bf x}_j)\}$ obtained by homography-based feature tracking, we estimate the camera pose ${\bf C}_i$ by minimizing both the relative pose error and re-projection error
\begin{equation}
	\small
	\begin{aligned}
		{\bf C}_i^* &= \argmin_{{\bf C}_i} ||\text{log}({\bf C}_i {\bf C}_{i-1}^{-1} {\bf T}_{(i-1)\rightarrow i}^{-1})||_{{\bm\Sigma}_{\bf T}}^2 + \\
		&\sum_{({\bf X}_j, {\bf x}_j) \in \mathcal{X}} \left(||\frac{{\bm\pi}({\bf K}({\bf C}_i {\bf X}_j)) - {\bf x}_j}{\sigma_{\bf x}}||_{\delta} + ||\frac{z^{-1}({\bf C}_i {\bf X}_j) - z^{-1}_j}{\sigma_z}||_{\delta}\right),\\
	\end{aligned}
	\label{eq:camera-pose-estimation}
	\normalsize
\end{equation}
where $\text{log}({\bf T})$ maps the 3D rigid transform ${\bf T} \in \text{SE}(3)$ to $\mathfrak{se}(3)$, and returns the minimal vector in $\mathbb{R}^6$. Here, $||{\bm\epsilon}||_{\bm\Sigma}^2 = {\bm\epsilon}^\top{\bm\Sigma}^{-1}{\bm\epsilon}$ is the squared Mahalanobis distance, and $z_j$ is the measured depth value at ${\bf x}_j$ in the current frame. $\sigma_{\bf x}$ and $\sigma_z$ are normalization parameters that control the corresponding weights. In our experiments, they are generally set to $1$ pixel and $0.05$, respectively.

\subsection{Depth Filtering}
If tracking fails, we invoke the relocalization procedure as in \cite{klein2008improving} to track features again. Otherwise, we use the new feature measurements to filter depths of those features.
In the pose estimation of (\ref{eq:camera-pose-estimation}), the 3D positions ${\bf X}$ are assumed to be known and kept fixed during optimization. ${\bf X}$ can be obtained directly from the depth measurement at the keyframe $k$ when the feature was first extracted, i.e. ${\bf X} = {\bm\pi}^{-1}({\bf C}_k, {\bf x}_k, z_k)$, where ${\bm\pi}^{-1}({\bf C}, {\bf x}, z) = {\bf C}^{-1}(z{\bf K}^{-1}\hat{\bf x})$. The depth value can be further refined by the following frames with new depth measurements. The depth filter must be robust to outliers since many features are extracted at object boundaries where depth measurements are unreliable.

We use the Bayesian depth filter proposed in \cite{VogiatzisH11}, which has been successfully used in other SLAM systems like SVO~\cite{forster2014svo} and REMODE~\cite{PizzoliFS14}. The filter continuously updates the joint distribution of the depth estimate $\rho$ and the inlier probability $\gamma$ by each incoming depth measurement $\hat\rho$ with variance $\hat\sigma^2$. The distribution of the measurement $\hat\rho$ given the correct $\rho$ and $\gamma$ is modeled as a \emph{Gaussian + uniform} mixture distribution:
\begin{equation}
P(\hat\rho|\rho,\gamma) = \gamma\mathcal{N}(\hat\rho|\rho,\hat\sigma^2) + (1-\gamma)\mathcal{U}(\hat\rho|\rho^\text{min},\rho^\text{max}).
\end{equation}
Given the set of measurements $\hat\rho_1,\cdots,\hat\rho_n$, the Bayesian posterior distribution of $\rho$ and $\gamma$ is estimated by
\begin{equation}\
\begin{aligned}
P(\rho,\gamma|\hat\rho_1,\cdots,\hat\rho_n) &\varpropto P(\rho,\gamma) \prod_i^n P(\hat\rho_i|\rho,\gamma)\\
&\varpropto P(\rho,\gamma|\hat\rho_1,\cdots,\hat\rho_{n-1}) P(\hat\rho_n|\rho,\gamma),\\
\end{aligned}
\label{eq:depth-posterior}
\end{equation}
where $P(\rho,\gamma)$ is the prior on $\rho$ and $\gamma$. Simply evaluating all probabilities of (\ref{eq:depth-posterior}) and choosing the best $\rho$ and $\gamma$ is computationally too
expensive. The authors of~\cite{VogiatzisH11} propose a parametric \emph{Gaussian} $\times$ \emph{beta} approximation to (\ref{eq:depth-posterior}):
\begin{equation}\
P(\rho,\gamma|\hat\rho_1,\cdots,\hat\rho_n) \approx \mathcal{N}(\rho|\mu_n,\sigma_n^2)\mathcal{B}(\gamma|a_n,b_n),
\label{eq:depth-posterior-approximation}
\end{equation}
where $\mu_n$, $\sigma_n^2$ are parameters controlling the Gaussian distribution, and $a_n$, $b_n$ are parameters controlling the beta distribution. These parameters are incrementally updated by each depth measurements. Substituting (\ref{eq:depth-posterior-approximation}) to (\ref{eq:depth-posterior}), we obtain
\begin{equation}\
\begin{aligned}
&\mathcal{N}(\rho|\mu_n,\sigma_n^2)\mathcal{B}(\gamma|a_n,b_n)\\
\varpropto &\mathcal{N}(\rho|\mu_{n-1},\sigma_{n-1}^2)\mathcal{B}(\gamma|a_{n-1},b_{n-1})P(\hat\rho_n|\rho,\gamma).\\
\end{aligned}
\label{eq:depth-filter-parameters}
\end{equation}
By matching the first and second order moments for $\rho$ and $\gamma$ in (\ref{eq:depth-filter-parameters}), the updated parameters $(\mu_n,\sigma_n^2,a_n,b_n)$ can be derived.

In our implementation, we use inverse depth~\cite{civera2008inverse}, which is better approximated by a Gaussian distribution, i.e. $\rho = 1/z$. 
For each feature correspondence $({\bf x}_k, {\bf x}_i)$ between keyframe $F_k$ and current frame $F_i$, we obtain a new measurement $\rho_{i\rightarrow k}$ for the inverse depth $\rho_k$ of ${\bf x}_k$ as:
\begin{equation}
\begin{aligned}
\rho_{i\rightarrow k} = \argmin_\rho &\frac{1}{\sigma_{\bf x}^2}||{\bm\pi}({\bf K}({\bf C}_i {\bm\pi}^{-1}({\bf C}_k, {\bf x}_k, \rho^{-1}))) - {\bf x}_i||_2^2\\
+ &\frac{1}{\sigma_z^2}(z^{-1}({\bf C}_i {\bm\pi}^{-1}({\bf C}_k, {\bf x}_k, \rho^{-1})) - z^{-1}_i)^2,\\
\end{aligned}
\end{equation}
where ${\bm\pi}^{-1}({\bf C}, {\bf x}, z) = {\bf C}^{-1}(z{\bf K}^{-1}\hat{\bf x})$. 
The filter continuously updates the joint distribution of $\rho_k$ and its inlier probability by each incoming $\rho_{i\rightarrow k}$. Please see~\cite{VogiatzisH11} for more details. At last, we decide whether $F_i$ is selected as a new keyframe. The keyframe which has maximal number of feature matches with $F_i$ is denoted as $F_{K_i}$. If the difference of view angle between $F_{K_i}$ and $F_i$ exceeds $45^\circ$, or the distance between them exceeds $0.5 z_{K_i}$~($z_{K_i}$ is the mean depth of $F_{K_i}$), then we select $F_i$ as a new keyframe.

\begin{figure}[t]
	\centering
	\includegraphics[width=1.0\linewidth]{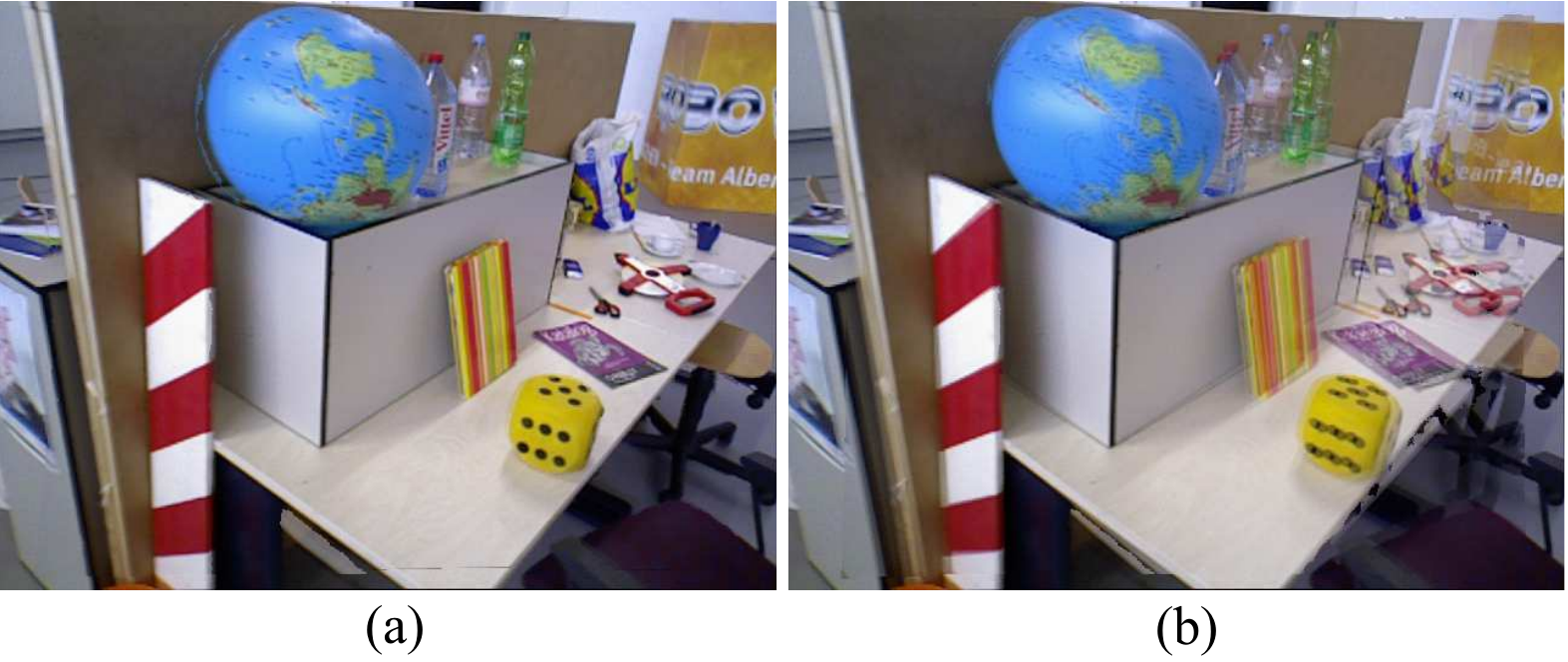}
	\caption{Comparison with/without low-resolution RGB-D alignment. (a) Warping frame 27 to frame 28 with the camera pose estimated by combining feature tracking and low resolution RGB-D alignment. (b) Warping frame 27 to frame 28 with the camera pose estimated by feature tracking only. (a) has much better alignment result than (b), which indicates that the estimated camera pose is more accurate.} \label{fig:alignment-comparison}
\end{figure}

Figure~\ref{fig:alignment-comparison} shows a comparison with and without low-resolution RGB-D alignment. For simulating fast motion, we extract one frame for
every 10 frames from ``fr3\_long\_office'' sequence in TUM RGB-D dataset~\cite{sturm12iros} to constitute a new sequence and then
perform tracking. With low-resolution RGB-D alignment, the tracking robustness is significantly improved but the computation does not increase much.

\section{Incremental Bundle Adjustment}

BA is performed when a new keyframe is inserted, or a new loop is found. In the former case, performing global BA seems to be unnecessary because only the local map will actually change. However, only performing local BA tends to be suboptimal especially when the local map contains large error. In that case it is better to involve more variables to BA, or else the error cannot be completely eliminated. We propose an efficient incremental BA~(called EIBA) that is able to provide nearly the same solution as global BA, but with significantly less computation time, which is proportional to how many variables are actually changed.

Before diving into our EIBA, we first revisit the algorithm of standard BA~\cite{Triggs99}. For easier illustration, we first introduce a regular BA function as follows:
\begin{equation}
	\sum_j\sum_{i\in\mathcal{V}_j} \left(||\frac{{\bm\pi}({\bf K}({\bf C}_i {\bf X}_j)) - {\bf x}_{ji}}{\sigma_{\bf x}}||_{\delta} + ||\frac{z^{-1}({\bf C}_i {\bf X}_j) - z^{-1}_{ji}}{\sigma_z}||_{\delta}\right),\nonumber
\end{equation}
which contains re-projection error term and inverse depth prior term. $\mathcal{V}_j$ is the set of cameras in which point $j$ is visible. The Huber norms can be converted to the form of squared sum using re-weighting scheme~\cite{hartley2003multiple}:
\begin{equation}
	f = \sum_j\sum_{i\in\mathcal{V}_j} ||{\bf f}_{ij}({\bf C}_i, {\bf X}_j)||_2^2,
\end{equation}
where ${\bf f}_{ij}\in\mathbb{R}^3$ (first two components for image re-projection and the third for depth prior). At each iteration, ${\bf f}_{ij}$ is linearized at the current estimate as
\begin{equation}
	{\bf f}_{ij}({\bf C}_i, {\bf X}_j) \approx {\bf J}_{{\bf C}_{ij}}\delta_{{\bf C}_i} + {\bf J}_{{\bf X}_{ij}}\delta_{{\bf X}_j} - {\bf e}_{ij},
	\label{eq:linearized-equation}
\end{equation}
where ${\bf J}_{{\bf C}_{ij}}$ is the Jacobian of ${\bf f}_{ij}$ with respect to ${\bf C}_i$, ${\bf J}_{{\bf X}_{ij}}$ is the Jacobian of ${\bf f}_{ij}$ with respect to ${\bf X}_j$, and ${\bf e}_{ij}$ is the residual error of ${\bf f}_{ij}$.
So we have
\begin{equation}
	f \approx ||{\bf J} {\bm\delta} - {\bf e}||_2^2,
\end{equation}
where ${\bf J}$ is the $3n_x\times(6n_c+3n_p)$ Jacobian matrix, ${\bf e}$ is the error vector, $n_x$ is the number of re-projection functions, $n_c$ and $n_p$ is the number of cameras and points respectively. ${\bm\delta}$ is the variable for the current iteration, ${\bm\delta} = [\delta_{\bf C}^\top, \delta_{\bf X}^\top]^\top$, $\delta_{\bf C} = [\delta_{{\bf C}_1}^\top,\cdots,\delta_{{\bf C}_{n_c}}^\top]^\top$ and $\delta_{\bf X} = [\delta_{{\bf X}_1}^\top,\cdots,\delta_{{\bf X}_{n_p}}^\top]^\top$. The update ${\bm\delta}$ is obtained by solving the normal equations
\begin{equation}
	{\bf J}^\top {\bf J} {\bm\delta} = {\bf J}^\top {\bf e}.
\end{equation}
Since each ${\bf f}_{ij}$ relates only one camera and one point, the normal equations are sparse and have the following form:
\begin{equation}
	\left[\begin{matrix}
		{\bf U}&{\bf W}\\
		{\bf W}^\top&{\bf V}
	\end{matrix}\right]
	\left[\begin{matrix}
		\delta_{\bf C}\\
		\delta_{\bf X}
	\end{matrix}\right]
	= \left[\begin{matrix}
		{\bf u}\\
		{\bf v}
	\end{matrix}\right],
	\label{eq:normal-equation}
\end{equation}
where ${\bf U}$ and ${\bf V}$ are $n_c\times n_c$ and $n_p\times n_p$ block matrices respectively, and only the diagonal block matrices ${\bf U}_{ii}$ and ${\bf V}_{jj}$ are non-zero. ${\bf W}$ is a $n_c\times n_p$ block matrix with non-zero block matrices ${\bf W}_{ij}$ if and only if point $j$ is visible in camera $i$. For efficient indexing and computation, we actually do not construct the whole matrices for ${\bf U}$, ${\bf V}$ and ${\bf W}$. Similar to \cite{jeong2012pushing}, we compute and store the small non-zero block matrices ${{\bf U}_{ii}}$, ${{\bf V}_{ii}}$, ${{\bf W}_{ii}}$. Compared to using general sparse matrix format, this data structure is more efficient and requires less memory space. Especially, when new keyframes or 3D points are added during incremental reconstruction, we do not need to reconstruct ${\bf J}^\top {\bf J}$ from scratch and only need to add new block matrices.

We first introduce the standard BA procedure as described in Algorithm~\ref{alg:standard-ba}. Equation (\ref{eq:normal-equation}) can be efficiently constructed as in step 1. A common strategy to solve (\ref{eq:normal-equation}) is to marginalize all points to construct the Schur complement and solve $\delta_{\bf C}$ first
\begin{equation}
	\begin{aligned}
		&{\bf S} \delta_{\bf C} = {\bf g},\\
		&{\bf S} = ({\bf U} - {\bf W}{\bf V}^{-1}{\bf W}^\top),\\
		&{\bf g} = {\bf u} - {\bf W}{\bf V}^{-1}{\bf v}.
	\end{aligned}
	\label{eq:schur-complement}
\end{equation}
Note that ${\bf S}$ is also sparse, with non-zero block matrix ${\bf S}_{i_1i_2}$ if and only if camera $i_1$ and $i_2$ share common points, thus can be efficiently constructed as in step 2 of Algorithm~\ref{alg:standard-ba}. The sparseness of ${\bf S}$ can also be exploited to solve $\delta_{\bf C}$. We use the preconditioned conjugate gradient (PCG) algorithm~\cite{jeong2012pushing} which naturally leverages the sparseness of ${\bf S}$. With solved $\delta_{\bf C}$, each $\delta_{{\bf X}_j}$ can be solved separately by back substitution:
\begin{equation}
	\delta_{{\bf X}_j} = {\bf V}_{jj}^{-1}\left({\bf v}_j - \sum_{i\in\mathcal{V}_j}{\bf W}_{ij}^\top\delta_{{\bf C}_i}\right).
	\label{eq:point-back-substitution}
\end{equation}

\begin{table} [tb!]
	\caption{Computational complexity for each step in standard BA (Algorithm~\ref{alg:standard-ba}).}
	\begin{center}
		\begin{tabular}{|c|c|}
			\hline Step &Complexity\\
			\hline 1&$O(Nl)$ or $O(Km)$\\
			\hline 2&$O(Nl^2)$ or $O(Kml)$\\
			\hline 3&$O(Kt)$\\
			\hline 4&$O(N)$ or $O(Km/l)$\\
			\hline
		\end{tabular}
	\end{center}
	\label{tab:complexity}
\end{table}

Because of the criterion of keyframe selection, only a small number of keyframe pairs share common points. We assume the number of 3D points is $N$, the number of keyframes is $K$, and the average observation number of each 3D point in keyframes is $l$. So the average number of the observations in each keyframe is $m=\frac{Nl}{K}$. We assume the average number of PCG iterations is $t$.
Table~\ref{tab:complexity} lists the computational complexity for each step, showing that most computation is required for steps 1 and 2. In most cases, $m$ is much larger than the number of PCG iterations $t$, typically hundreds for $m$ and dozens for $t$. Then the computation consumed in steps 1 and 2 would be hundreds of times larger than for step 3.

During incremental reconstruction, most variables are nearly unchanged after global BA, thus most computation in steps 1, 2, and 4 are actually unnecessary. Specifically, in step 1, the contribution of most ${\bf f}_{ij}$s to (\ref{eq:normal-equation}) nearly remains the same between successive iterations. Here, we propose an efficient incremental BA~(EIBA) algorithm which can make maximum use of intermediate computation to save computation. As shown in Algorithm~\ref{alg:incremental-ba}, instead of constructing (\ref{eq:normal-equation}) from scratch at each iteration, we update (\ref{eq:normal-equation}) from the last iteration. We store the effect of ${\bf f}_{ij}$ to (\ref{eq:normal-equation}) in ${\bf A}^{\bf U}_{ij}$, ${\bf A}^{\bf V}_{ij}$, ${\bf b}^{\bf u}_{ij}$ and ${\bf b}^{\bf v}_{ij}$. We initialize ${\bf A}^{\bf U}_{ij}=0$, ${\bf A}^{\bf V}_{ij}=0$, ${\bf b}^{\bf u}_{ij}=0$ and ${\bf b}^{\bf v}_{ij}=0$ in the beginning. They are re-computed if and only if the linearization point of ${\bf f}_{ij}$ is changed. In this case, we remove their contribution to (\ref{eq:normal-equation}) from the last iteration, refresh them, and update (\ref{eq:normal-equation}) for the current iteration. If and only if ${\bf V}_{jj}$ is updated, point $j$ must be re-marginalized. Then we update point marginalization and Schur complement (\ref{eq:schur-complement}) in a similar way, see Algorithm~\ref{alg:incremental-ba} for details. In step 3, we solve (\ref{eq:schur-complement}) by PCG, and change ${\bf C}_i$ only if $||\delta_{{\bf C}_i}||$ exceeds a threshold $\epsilon_c$. In step 4, we perform back substitution only for points visible in the changed cameras, and change ${\bf X}_j$ only if $||\delta_{{\bf X}_j}||$ exceeds a threshold $\epsilon_p$.

\begin{algorithm}
	\caption{One iteration in standard BA}
	\begin{enumerate}
		\item {\small Construct normal equations (\ref{eq:normal-equation})}
		\begin{algorithmic}
			\State {\small ${\bf U} = {\bf 0}$; ${\bf V} = {\bf 0}$; ${\bf W} = {\bf 0}$; ${\bf u} = {\bf 0}$; ${\bf v} = {\bf 0}$}
			\For{{\small each point $j$ and each camera $i\in\mathcal{V}_j$}}
			\State {\small Construct linearized equation (\ref{eq:linearized-equation})}
			\State {\small ${\bf U}_{ii} += {\bf J}_{{\bf C}_{ij}}^\top{\bf J}_{{\bf C}_{ij}}$}
			\State {\small ${\bf V}_{jj} += {\bf J}_{{\bf X}_{ij}}^\top{\bf J}_{{\bf X}_{ij}}$}
			\State {\small ${\bf u}_i += {\bf J}_{{\bf C}_{ij}}^\top{\bf e}_{ij}$}
			\State {\small ${\bf v}_j += {\bf J}_{{\bf X}_{ij}}^\top{\bf e}_{ij}$}
			\State {\small ${\bf W}_{ij} = {\bf J}_{{\bf C}_{ij}}^\top{\bf J}_{{\bf X}_{ij}}$}
			\EndFor
		\end{algorithmic}
		\item {\small Marginalize points to construct Schur complement (\ref{eq:schur-complement})}
		\begin{algorithmic}
			\State {\small ${\bf S} = {\bf U}$}
			\For{{\small each point $j$ and each camera pair $(i_1,i_2)\in\mathcal{V}_j\times\mathcal{V}_j$}}
			\State {\small ${\bf S}_{i_1i_2} -= {\bf W}_{i_1j}{\bf V}_{jj}^{-1}{\bf W}_{i_2j}^\top$}
			\EndFor
			\State {\small ${\bf g} = {\bf u}$}
			\For{{\small each point $j$ and each camera $i\in\mathcal{V}_j$}}
			\State {\small ${\bf g}_i -= {\bf W}_{ij}{\bf V}_{jj}^{-1}{\bf v}_j$}
			\EndFor
		\end{algorithmic}
		\item {\small Update cameras}
		\begin{algorithmic}
			\State {\small Solve $\delta_{{\bf C}_i}$ in (\ref{eq:schur-complement}) using PCG~\cite{jeong2012pushing}}
			\For{{\small each keyframe $i$}}
			\State {\small ${\bf C}_i = \text{exp}(\delta_{{\bf C}_i}){\bf C}_i$}
			\EndFor
		\end{algorithmic}
		\item {\small Update points}
		\begin{algorithmic}
			\For{{\small each point $j$}}
			\State {\small Solve $\delta_{{\bf X}_j}$ by (\ref{eq:point-back-substitution})}
			\State {\small ${\bf X}_j += \delta_{{\bf X}_j}$}
			\EndFor
		\end{algorithmic}
	\end{enumerate}
	\label{alg:standard-ba}
\end{algorithm}

\begin{algorithm}
	\caption{One iteration in our incremental BA}
	\begin{enumerate}
		\item {\small Update normal equations (\ref{eq:normal-equation}) and Schur complement (\ref{eq:schur-complement})}
		\begin{algorithmic}
			\For{{\small each point $j$ and each camera $i\in\mathcal{V}_j$ that ${\bf C}_i$ or ${\bf X}_j$ is changed}}
			\State {\small Construct linearized equation (\ref{eq:linearized-equation})}
			\State {\small ${\bf S}_{ii} -= {\bf A}^{\bf U}_{ij}$; ${\bf A}^{\bf U}_{ij} = {\bf J}_{{\bf C}_{ij}}^\top{\bf J}_{{\bf C}_{ij}}$; ${\bf S}_{ii} += {\bf A}^{\bf U}_{ij}$}
			\State {\small ${\bf V}_{jj} -= {\bf A}^{\bf V}_{ij}$; ${\bf A}^{\bf V}_{ij}={\bf J}_{{\bf X}_{ij}}^\top{\bf J}_{{\bf X}_{ij}}$; ${\bf V}_{jj} += {\bf A}^{\bf V}_{ij}$}
			\State {\small ${\bf g}_i -= {\bf b}^{\bf u}_{ij}$; ${\bf b}^{\bf u}_{ij} = {\bf J}_{{\bf C}_{ij}}^\top{\bf e}_{ij}$; ${\bf g}_i += {\bf b}^{\bf u}_{ij}$}
			\State {\small ${\bf v}_j -= {\bf b}^{\bf v}_{ij}$; ${\bf b}^{\bf v}_{ij} = {\bf J}_{{\bf X}_{ij}}^\top{\bf e}_{ij}$; ${\bf v}_j += {\bf b}^{\bf v}_{ij}$}
			\State {\small ${\bf W}_{ij} = {\bf J}_{{\bf C}_{ij}}^\top{\bf J}_{{\bf X}_{ij}}$}
			\State {\small Mark ${\bf V}_{jj}$ updated}
			\EndFor
		\end{algorithmic}
		\item {\small Update point marginalization and Schur complement (\ref{eq:schur-complement})}
		\begin{algorithmic}
			\For{{\small each point $j$ that ${\bf V}_{jj}$ is updated and each camera pair $(i_1,i_2)\in\mathcal{V}_j\times\mathcal{V}_j$}}
			\State {\small ${\bf S}_{i_1i_2} += {\bf A}^{\bf S}_{i_1i_2j}$}
			\State {\small ${\bf A}^{\bf S}_{i_1i_2j} = {\bf W}_{i_1j}{\bf V}_{jj}^{-1}{\bf W}_{i_2j}^\top$}
			\State {\small ${\bf S}_{i_1i_2} -= {\bf A}^{\bf S}_{i_1i_2j}$}
			\EndFor
			\For{{\small each point $j$ that ${\bf V}_{jj}$ is updated and each camera $i\in\mathcal{V}_j$}}
			\State {\small ${\bf g}_i += {\bf b}^{\bf g}_{ij}$; ${\bf b}^{\bf g}_{ij} = {\bf W}_{ij}{\bf V}_{jj}^{-1}{\bf v}_j$; ${\bf g}_i -= {\bf b}^{\bf g}_{ij}$}
			\EndFor
		\end{algorithmic}
		\item {\small Update cameras}
		\begin{algorithmic}
			\State {\small Solve $\delta_{{\bf C}_i}$ in (\ref{eq:schur-complement}) using PCG~\cite{jeong2012pushing}}
			\For{{\small each keyframe $i$ that $||\delta_{{\bf C}_i}||>\epsilon_c$}}
			\State {\small ${\bf C}_i = \text{exp}(\delta_{{\bf C}_i}){\bf C}_i$}
			\State {\small Mark ${\bf C}_i$ changed}
			\EndFor
		\end{algorithmic}
		\item {\small Update points}
		\begin{algorithmic}
			\For{{\small each point $j$ that any ${\bf C}_i$ with $i\in\mathcal{V}_j$ is changed}}
			\State {\small Solve $\delta_{{\bf X}_j}$ by (\ref{eq:point-back-substitution})}
			\If {\small $||\delta_{{\bf X}_j}||>\epsilon_p$}
			\State {\small ${\bf X}_j += \delta_{{\bf X}_j}$}
			\State {\small Mark ${\bf X}_j$ changed}
			\EndIf
			\EndFor
		\end{algorithmic}
	\end{enumerate}
	\label{alg:incremental-ba}
\end{algorithm}

The above paragraphs introduce the incremental optimization with a regular BA function.
Actually, our EIBA is quite general and can be naturally extended to solve the following energy function:
\begin{equation}
\begin{aligned}
\sum_j\sum_{i\in\mathcal{V}_j} &\left(||\frac{{\bm\pi}({\bf K}({\bf C}_i {\bf X}_j)) - {\bf x}_{ji}}{\sigma_{\bf x}}||_{\delta} + ||\frac{z^{-1}({\bf C}_i {\bf X}_j) - z^{-1}_{ji}}{\sigma_z}||_{\delta}\right)\\
+ &\sum_{(i_1,i_2)\in\mathcal{L}}||\text{log}({\bf C}_{i_1}\circ{\bf C}_{i_2}\circ{\bf T}_{i_1i_2}^{-1})||^2_{{\bf\Sigma}_{i_1i_2}},\\
\end{aligned}
\label{eq:ourBA}
\end{equation}
where $\mathcal{L}$ is the set of loop constraint. Each loop constraint is represented as relative pose ${\bf T}_{i_1i_2}$ with covariance ${\bm\Sigma}_{i_1i_2}$. It does not harm the sparseness of normal equation or Schur complement. When the state of ${\bf C}_{i_1}$ or ${\bf C}_{i_2}$ is changed, the error function is re-linearized as
\begin{equation}
{\bf f}({\bf C}_{i_1},{\bf C}_{i_2}) \approx {\bf J}_{i_1}\delta_{{\bf C}_{i_1}} + {\bf J}_{i_2}\delta_{{\bf C}_{i_2}} - {\bf e}.
\end{equation}
Then ${\bf J}_{i_1}^\top{\bf J}_{i_1}$, ${\bf J}_{i_1}^\top{\bf J}_{i_2}$ and ${\bf J}_{i_2}^\top{\bf J}_{i_2}$ are updated to ${\bf S}_{i_1i_1}$, ${\bf S}_{i_1i_2}$ and ${\bf S}_{i_2i_2}$ respectively. Similarly, ${\bf J}_{i_1}^\top{\bf e}$ and ${\bf J}_{i_2}^\top{\bf e}$ are updated to ${\bf g}_{i_1}$ and ${\bf g}_{i_2}$ respectively. In addition, we use inverse depth to parameterize ${\bf X}_j$. Assuming the first keyframe that ${\bf X}_j$ is observed is frame $k$, we have ${\bf X}_j={\bf C}^{-1}_k(z_{jk}{\bf K}^{-1}\hat{\bf x}_{jk})$. So each re-projection equation ${\bf f}_{ij}$ actually relates two camera poses~(i.e. ${\bf C}_i$ and ${\bf C}_k$) and one 3D point ${\bf X}_j$. So at each iteration, ${\bf f}_{ij}$ is linearized as
\begin{equation}
{\bf f}_{ij}({\bf C}_i, {\bf C}_k, {\bf X}_j) \approx {\bf J}_{{\bf C}_{ij}}\delta_{{\bf C}_i} + {\bf J}_{{\bf C}_{kj}}\delta_{{\bf C}_k} + {\bf J}_{{\bf X}_{ij}}\delta_{{\bf X}_j} - {\bf e}_{ij},
\label{eq:linearized-equation-inversedepth}
\end{equation}
where ${\bf J}_{{\bf C}_{kj}}$ is the Jacobian of ${\bf f}_{ij}$ with respect to ${\bf C}_k$. So in Step 1 of Algorithm~\ref{alg:incremental-ba}, we also need to update ${\bf S}_{kk}$, ${\bf S}_{ik}$, ${\bf W}_{kj}$ and ${\bf g}_k$ for each observation.

\begin{table*} [t!]
	{
		\caption{Timing comparison for incremental BA.}
		\begin{center}
			\begin{tabular}{|c|c|c|c|c|c|c|}
				\hline  Sequence & Num. of Camera / Points & Num. of Observations & EIBA & \multicolumn{3}{c|}{iSAM2 } \\
				  & & & & No relinearization & relinearizeSkip = 10 & relinearizeSkip = 5\\
				\hline  fr3\_long\_office & 92 / 4322 & 12027 & 88.9ms & 983.9ms & 1968.2ms & 2670.9ms\\
				\hline  fr2\_desk & 63 / 2780 & 6897 & 34.8ms & 507.8ms & 850.4ms & 1152.0ms\\
				\hline
			\end{tabular}
		\end{center}
		\label{tab:time-IBA}}
\end{table*}

Although our incremental Schur complement is essentially similar to \cite{IlaPSI17}, our computation is more efficient with block-wise matrix computation and storing the updated matrix blocks for reducing calculation. In addition, we use preconditioned conjugate gradient (PCG) algorithm to solve the linear system, which is more efficient than using factorization methods due to the following reasons: 1) block-based PCG can better leverage the sparseness to efficiently solve Schur complement as verified in \cite{jeong2012pushing}, and 2) PCG can make the most of the incremental nature because good initial values of the variables can be easily obtained in incremental BA so that a few iterations are generally enough to converge.

Kaess~\etal~\cite{kaess2011isam2} also performs incremental BA~(called iSAM2) by updating a matrix factorization. In iSAM2, the square root information matrix of ${\bf J}$ is encoded in a Bayes tree, in which each node contains a set of frontal variables $\mathcal{F}$ and represents a conditional density $P(\mathcal{F}|\mathcal{S})$, where $\mathcal{S}$ is contained in the frontal variables of its parent node. Inserting a new keyframe will only affect nodes containing its visible points as frontal variables and their ancestor nodes. All the affected variables will be re-eliminated. For efficiency, it is better to push the visible points of the new keyframe to the root, i.e. marginalizing these points in the end. However, if their visible cameras are marginalized before them, correlations among these points will occur, which significantly degrades efficiency. In addition, if the camera is moving to and fro, a large number of invisible points may also be affected by iSAM2. By comparison, in our EIBA, points are always marginalized first to minimize fill-in, and only the points visible in the new keyframe will be affected. Although all the cameras will also be affected in step 3, since we are dealing with moderate size scenes, the number of cameras is much smaller than the number of points potentially affected by iSAM2. Besides, the Schur complement is very sparse, so that PCG is able to solve it very efficiently. 

\begin{figure}
	\includegraphics[width=1.0\linewidth]{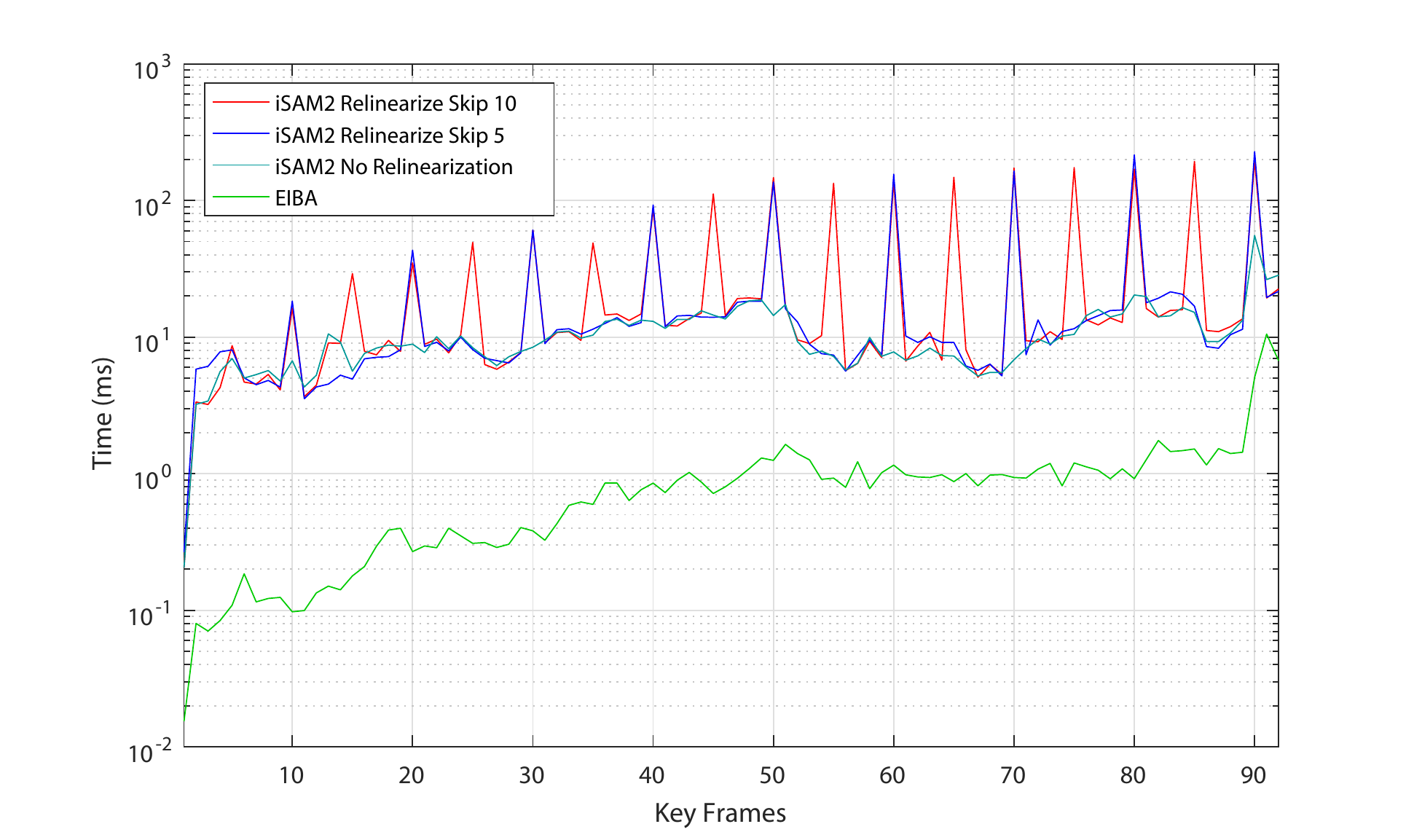}
	\caption{The computation time of our EIBA and iSAM2 while incrementally adding each new keyframe on ``fr3\_long\_office'' sequence.}
	\label{fig:time-ba}
\end{figure}

\begin{figure}
	\includegraphics[width=1.0\linewidth]{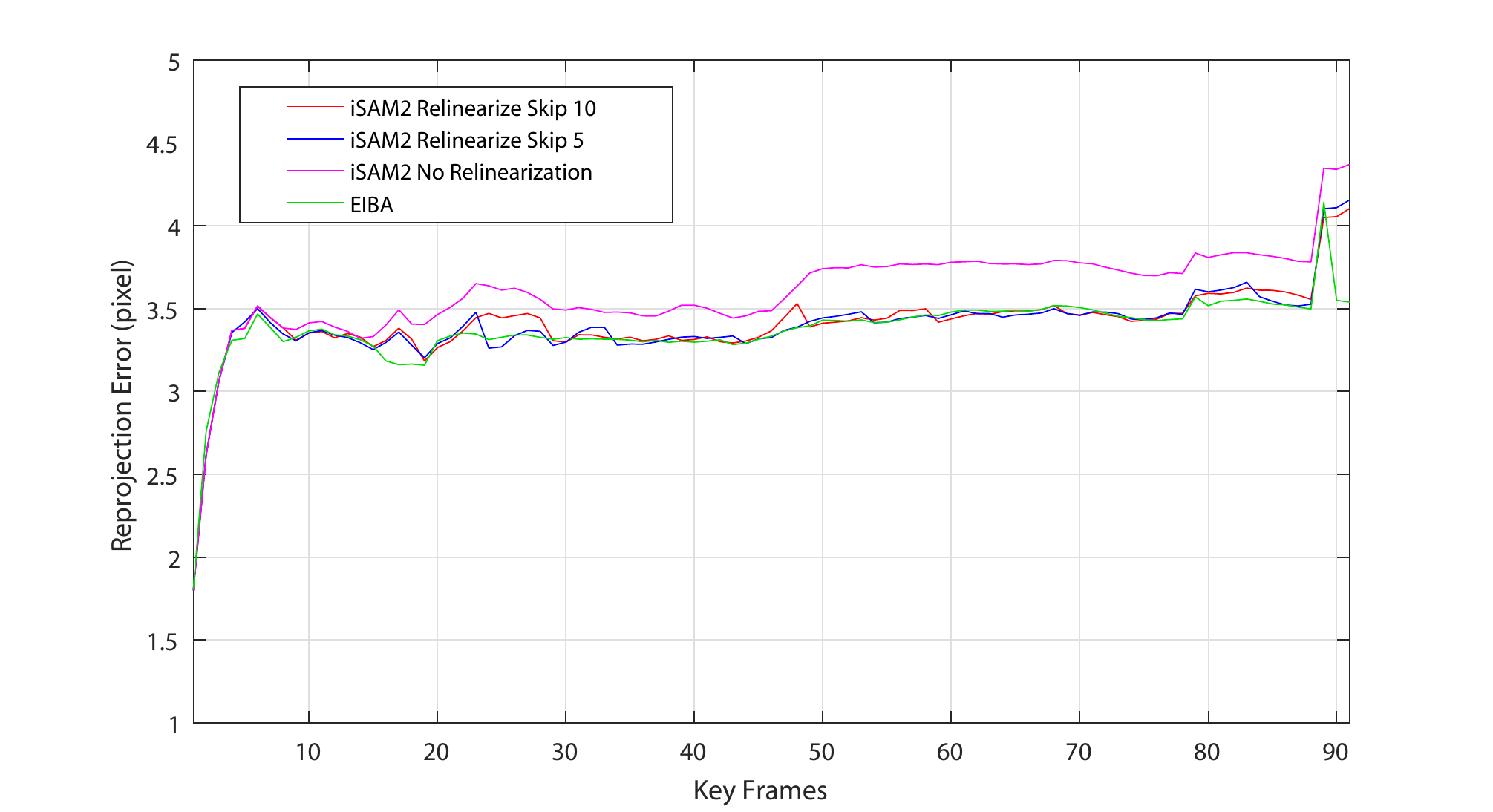}
	\caption{The optimized reprojection error~(RMSE) for our EIBA and iSAM2 while incrementally adding each new keyframe on ``fr3\_long\_office'' sequence.}
	\label{fig:error-ba}
\end{figure}

We use ``fr3\_long\_office'' and ``fr2\_desk'' sequences from TUM RGB-D benchmark~\cite{sturm12iros} to make comparisons with iSAM2. On ``fr3\_long\_office'', there are $92$ keyframes, $4,322$ 3D points and $12,027$ observations. On ``fr2\_desk'' sequence, there are $63$ keyframes, $2,780$ 3D points and $6,897$ observations. We perform incremental BA for (\ref{eq:ourBA}) while adding each new keyframe. For fair comparison, both two mehods perform only one iteration. Since our EIBA applies SSE instructions for code optimization, we also enable SSE optimization while compiling Eigen library used in iSAM2. We use Gauss-Newton optimization method for both EIBA and iSAM2. In addition, the linearization parameter will significantly influence the speed and accuracy of iSAM2. In our experiments, we test three configurations for iSAM2, i.e. no linearization, linearization every 10 steps, and linearization every 5 steps. The running time is tested on a desktop PC with i7 3.6GHz CPU and 16G memory. Figure~\ref{fig:time-ba} shows the computation time for EIBA and iSAM2 while adding each new keyframe~(the time of loading data is not included). Table~\ref{fig:time-ba} shows the total running time. As can be seen, for our EIBA, the computation almost keeps constant~(about $1$ms) when the number of keyframes increases, except when loop closure is detected, in which case the poses of more keyframes need to be updated. Although the computation of iSAM2 also almost keeps constant, it is generally slower than ours by an order of magnitude even without using linearization. The optimized reprojection error by EIBA is lower than iSAM2 without linearization and comparable with iSAM2 with relinearizeSkip = 5, as shown in Figure~\ref{fig:error-ba}. 

Especially, we found that the computation time of iSAM2 significantly increases when loop clousre is detected. In contrast, the computation time of EIBA does not increase so much (only increases to about $10 ms$). The reason is that when a large loop detected, the information matrix will become very dense and almost all varaibles need to be updated. In EBA, although the Schur complement matrix $\bf S$ will become very dense in this case, we use block-based PCG algorithm to solve the linear system which is more efficient than using a factoriation method.

\section{Keyframe-based Dense Mapping}
Similar to~\cite{MaierSC17}, we also do integration and de-integration in keyframes. We use a volumetric method with spatial hashing~\cite{NiessnerZIS13},  to fuse the depth maps to construct the complete 3D model. 
Kahler~\etal~\cite{InfiniTAM_ISMAR_2015} proposed a very efficient volumetric integration method based on voxel block hashing. We adapt this method for fast volumetric fusion.
When the camera pose of the current frame is estimated with good quality, we need to fuse the depth map into the global model. However, if we directly integrate the depth map of each frame and discard the frame, we could not correct the model again when a loop closure is detected or the poses of frames are refined by BA. A simple solution is to store all the depth maps and re-integrate them once the camera poses are refined. However, this will be intractable for real-time application since the number of frames always increases and may become very large. So we proposed to use keyframes to represent the depth data and operate integration and de-integration in keyframes.

If the current frame $F_i$ is selected as keyframe, we can directly integrate the depth map $D_i$ into the global model. For each voxel $\bf v$, its truncated signed distance is denoted as ${\bf D}({\bf v})$, and the weight is denoted as ${\bf W}({\bf v})$. For pixel $\bf x$ in $F_i$, its SDF is defined as $\phi({\bf x})=D_i({\bf x})-z_i({\bf v})$, where $z_i({\bf v})$ denotes the projected depth in $F_i$ for voxel $\bf v$. If $\phi({\bf x}) \geq -\mu$ where $\mu$ is a pre-defined truncated value, we can update the corresponding TSDF of $\bf v$ as
\begin{equation}
	\footnotesize
	{\bf D}'({\bf v})=\frac{{\bf D}({\bf v}){\bf W}({\bf v})+w_i({\bf x})\min(\mu,\phi({\bf x}))}{{\bf W}({\bf v}) + w_i({\bf x})},\\
	{\bf W}'({\bf v}) = {\bf W}({\bf v}) + w_i({\bf x}),
	\label{eq:integration}
\end{equation}
\normalsize
where $w_i({\bf x})$ is the integration weight for $\bf x$.

If $F_i$ is not selected as keyframe, we first find the keyframe which has maximal number of feature matches with $F_i$, denoted as $F_{K_i}$. We de-integrate the depth map of $F_{K_i}$ from the global model. Inspired by \cite{dai2017bundle}, the de-integration operation is similar to integration. If $\phi({\bf x}) \geq -\mu$, each voxel $\bf v$ can be updated as
\begin{equation}
	\footnotesize
	{\bf D}'({\bf v})=\frac{{\bf D}({\bf v}){\bf W}({\bf v}) - w_i({\bf x})\min(\mu,\phi({\bf x}))}{{\bf W}({\bf v}) - w_i({\bf x})},
	{\bf W}'({\bf v}) = {\bf W}({\bf v}) - w_i({\bf x}).
	\label{eq:de-integration}
\end{equation}
\normalsize

After de-integration, we fuse depth map $D_i$ into $F_{K_i}$ by projecting it to $F_{K_i}$, which is similar to that in \cite{MaierSC17}. The major difference is that we take into account the occlusion and store the unfused depths instead of simply discarding. For pixel $\bf x$ in $F_i$, its projection in $F_{K_i}$ is denoted as $\bf y$. If the difference of the inverse depth of pixel $\bf y$~(i.e. $1/D_{F_{K_i}}({\bf y})$) and the projected inverse depth of $\bf x$~(denoted as $1/z^{i \to K_i}_{\bf x}$) is less than a threshold $\tau_d$, we filter the depth of $\bf y$ as
\begin{equation}
	\footnotesize
	D'({\bf y}) = \frac{w_{K_i}({\bf y}) D_{K_i}({\bf y}) + w_i({\bf x})z^{i \to K_i}_{\bf x}}{w_{K_i}({\bf y}) + w_i({\bf x})}, 
	w_{K_i} = w_{K_i}({\bf y}) + w_i({\bf x}).
	\label{eq:filter}
\end{equation}
About $w_i({\bf x})$ in (\ref{eq:integration}), (\ref{eq:de-integration}) and (\ref{eq:filter}), we set it as follows: if $\bf x$ is in a key frame, $w_i(x)$ is set to the filtering number of $\bf x$, otherwise it is set to 1.

We count the fusion number $N_{K_i}$ for each keyframe $F_{K_i}$ to control the maximum number of depth fusion since too many fusions may be unnecessary and even degrade the reconstruction quality.
Since the overlap of $F_i$ and $F_{K_i}$ is generally large, most depths of the current frame can be fused into $F_{K_i}$ except some pixels that are occluded or out of view. This strategy can significantly reduce the depth redundancy. If the number of unfused depths is less than a threshold $\tau$, we simply discard these unfused 3D points. Otherwise, we create a point cloud set $V_i$ to store these unfused 3D points, and link it to $F_{K_i}$~(we store the relative pose between $F_i$ and $F_{K_i}$ for 3D points projection during integration and de-integration). Then we integrate the updated depth map $D_{K_i}$. If $V_i$ is not empty, the 3D points in $V_i$ are also projected into $F_i$, and then we perform integration on $F_i$. So for each incoming frame that is not selected as keyframe, we perform two integrations and one de-integration. Since the number of unfused 3D points in $F_i$ is small, the integration time is also small. So the computation time of our keyframe-based fusion is generally slightly larger than two times that of the traditional volumetric fusion method. In \cite{MaierSC17}, they first fuse a constant number of non-keyframes to a nearest keyframe, and then integrate the depth map of this keyframe to the 3D model. The 3D model will be not updated until the keyframe fusion is finished. In our method, since we first de-integrate the old depth map of keyframe $F_{K_i}$ from 3D model and then re-integrate the updated depth map immediately while fusing each non-keyframe $F_i$ to $F_{K_i}$, the 3D model can be timely updated without delay.
If $N_{K_i}$ is large and the number of keyframes is not increased for a long time, it means that not sufficient new content has been scanned. In this case, we simply discard $D_i$ without fusion.

\begin{figure}[tb]
	\centering
	\includegraphics[width=1.0\linewidth]{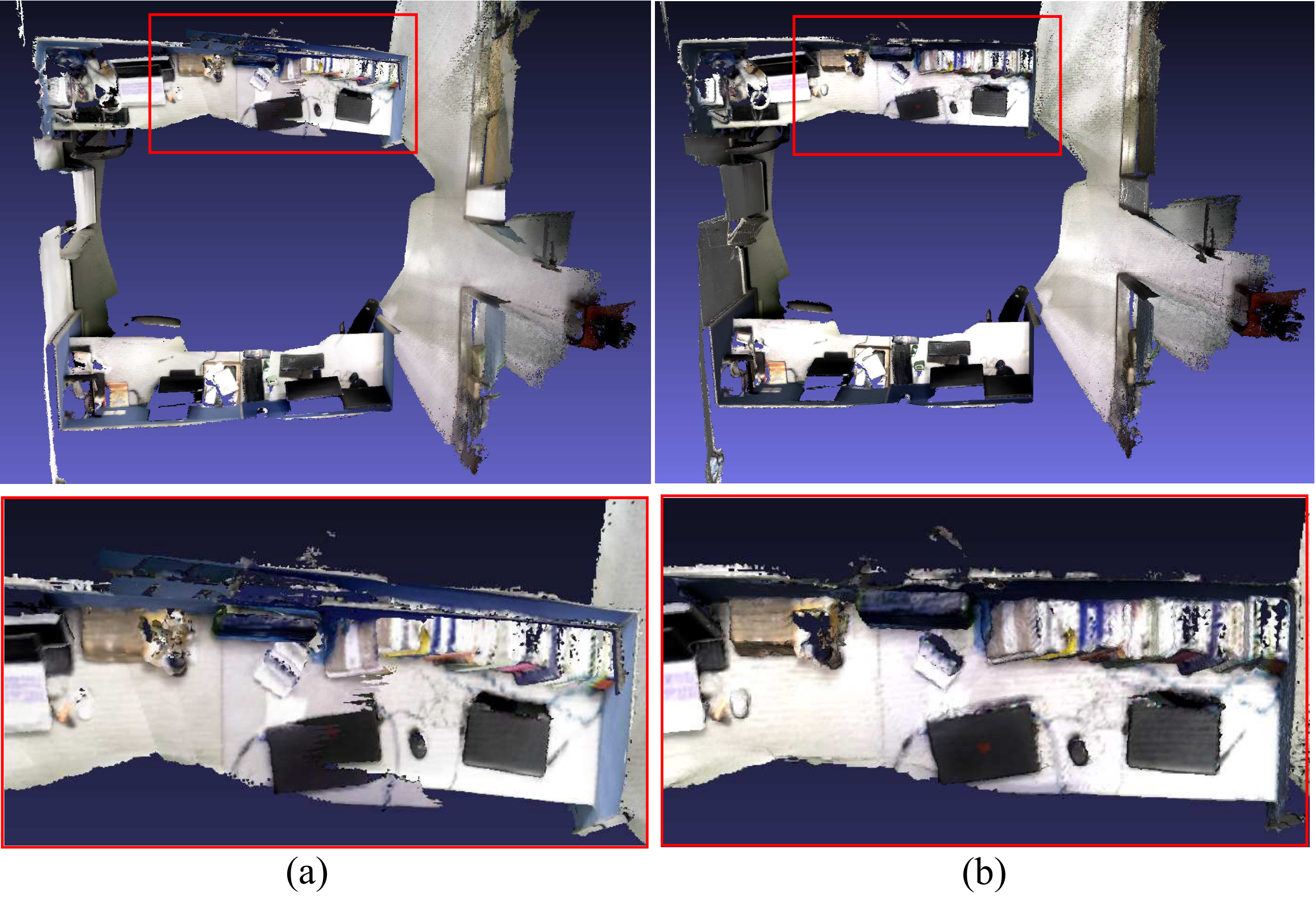}
	\caption{Comparison with/without re-integration. (a) The reconstructed 3D model without re-integration. (b) The reconstructed 3D model with our keyframe-based re-integration, which is more accurate and globally consistent than (a), as highlighted with the red rectangle.} \label{fig:fusion-comparison}
\end{figure}

To further reduce redundancy, if the current frame $F_i$ is selected as keyframe, we find the point cloud sets linked to nearby keyframes and fuse them to $F_i$. If the remaining number of points of $V_j$ is too small after fusion~(less than $2,000$ in our experiments), we will simply discard it and de-integrate it from the global map. Otherwise, we only de-integrate the 3D points which have been fused into $F_i$. For real-time computation, we only fuse $5\sim10$ point cloud sets in our experiments.

If the poses of keyframes are modified by BA, we need to re-integrate the depth maps of all the updated keyframes and their linked point cloud sets. However, if the number of adjusted keyframes is large, the re-integration time will be too large to satisfy real-time applications. Therefore, we propose to limit the number of re-integration operations for each time instance. We maintain a update queue for the keyframes which poses have been updated. The keyframes with largely changed poses will be re-integrated with higher priority. This strategy can guarantee that the mapping can always run with almost constant speed even when BA is invoked. In \cite{MaierSC17}, they perform depth re-integration only when receiving a pose update, and the uncorrected depth maps of the adjusted poses need to wait for re-integration in a final pass. In contrast, we perform surface update for each time instance so that the surface can be corrected more timely. The uncorrected depth maps of the updated poses still have chance to be re-integrated in the next time instance, and do not need to wait for the re-integration in a final pass. Figure~\ref{fig:fusion-comparison} shows a comparison with and without re-integration. As can be seen, due to accumulation error, if we do not re-integrate the depth maps, the reconstructed 3D surface has obvious artifacts. In contrast, with our keyframe-based re-integration, the reconstructed 3D surface becomes more accurate and globally consistent.

For further acceleration, we can fuse only one frame out of every two or more frames, which does not degrade much the reconstruction quality but can significantly accelerate the volumetric fusion.

\begin{figure}
	\includegraphics[width=1.0\linewidth]{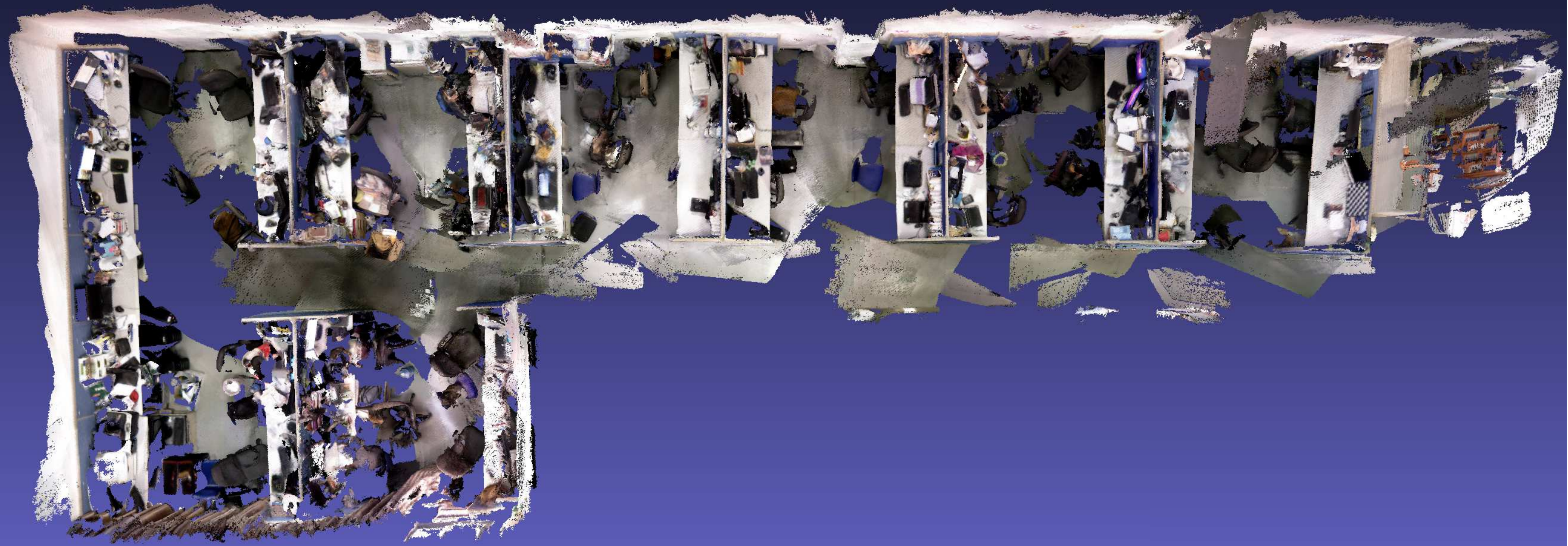}
	\caption{Our reconstructed 3D model for ``Cubes'' sequence.}
	\label{fig:loop-closure}
\end{figure}

\begin{figure}
	\includegraphics[width=1.0\linewidth]{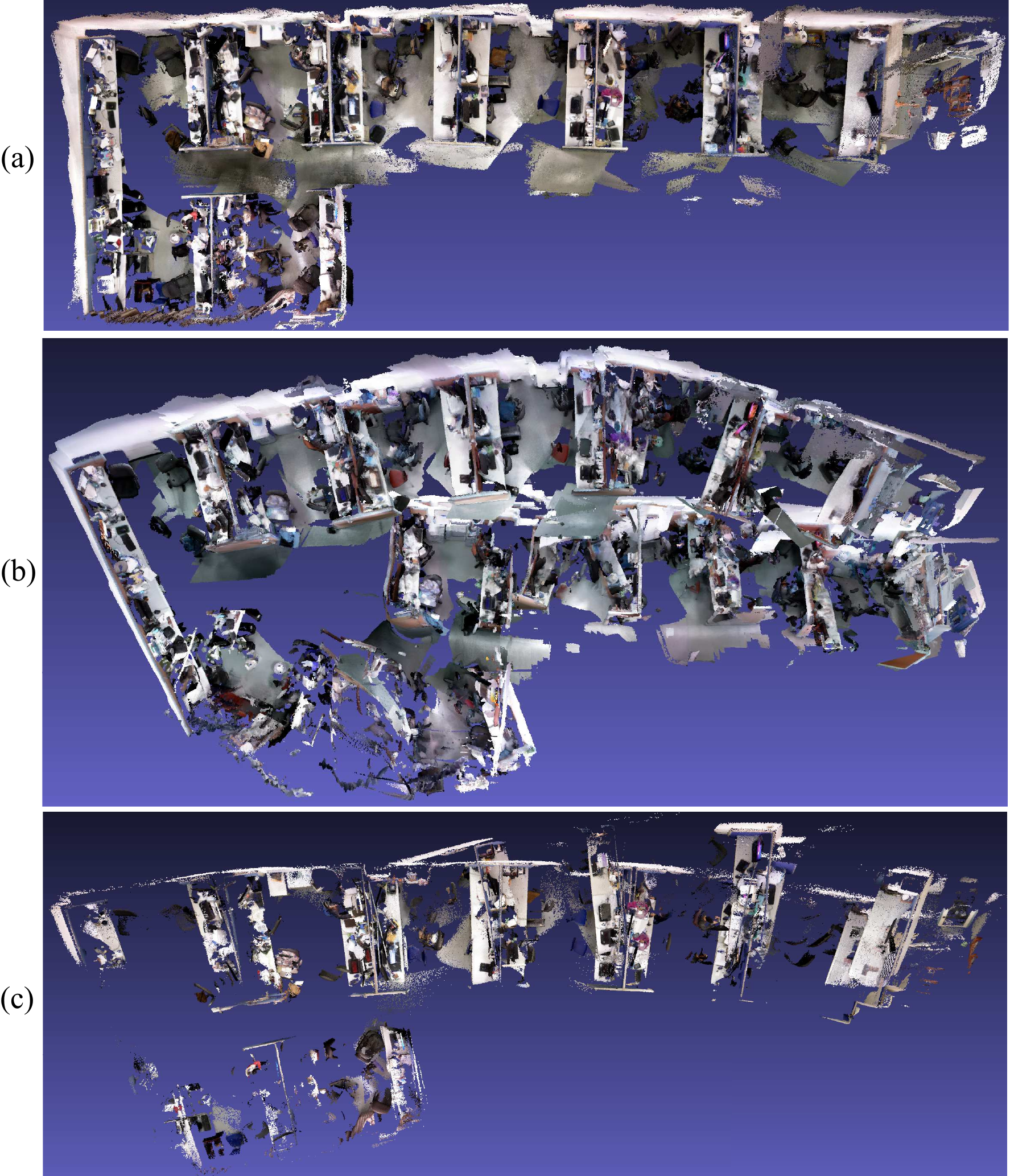}
	\caption{Comparison in a low-frame-rate example. (a) The reconstruction of ours. (b) The reconstruction of Kintinuous. (c) The reconstruction of ElasticFusion.}
	\label{fig:low-frame-rate}
\end{figure}

\begin{figure}
	\includegraphics[width=1.0\linewidth]{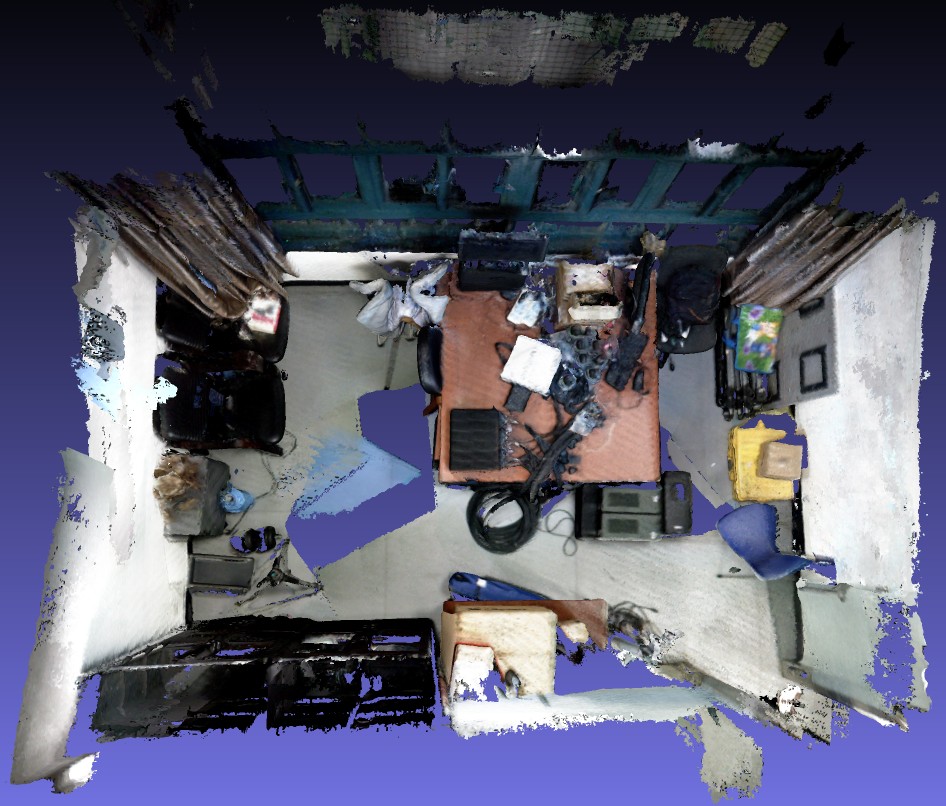}
	\caption{Our reconstruction result of ``Office'' dataset with $20,862$ frames in total.}
	\label{fig:long-time}
\end{figure}

\section{Experimental Results}
We have conducted experiments with both TUM RGB-D benchmark~\cite{sturm12iros} and indoor sequences captured by ourselves.
On a desktop PC with an Intel i5 3.3GHz CPU, 20GB memory and GTX 1070 graphics card~(8GB video memory), the tracking component without GPU acceleration takes about $5\sim14$ms per frame, and the dense mapping component in the foreground thread takes about $1.2\sim6$ms per frame. The whole system enabling both tracking and dense mapping runs above 50fps. For a laptop with an Intel i7 2.6GHz CPU, 16GB memory and GTX 960M graphics card~(4GB video memory), the system runs around 30fps. If we fuse only one out of three frames, the running time could be even faster.

\begin{table*} [tb!]
	{
		\caption{Comparison of ATE RMSE on all of the scenes in the TUM RGB-D benchmark.}
		\begin{center}
			\footnotesize
			\begin{tabular}{|@{\hspace{1mm}}c@{\hspace{1mm}}|@{\hspace{1mm}}c@{\hspace{1mm}}|@{\hspace{1mm}}c@{\hspace{1mm}}|@{\hspace{1mm}}c@{\hspace{1mm}}|@{\hspace{1mm}}c@{\hspace{1mm}}|@{\hspace{1mm}}c@{\hspace{1mm}}|@{\hspace{1mm}}c@{\hspace{1mm}}|@{\hspace{1mm}}c@{\hspace{1mm}}|@{\hspace{1mm}}c@{\hspace{1mm}}|}
				\hline   & Ours & Ours & Kintinuous & ElasticFusion & DVO-SLAM & RGB-D SLAM & MRSMap & BundleFusion\\
				& (all frames) & (key frames) & & & & & & \\
				\hline  fr1\_360  & 13.0cm & 10.9cm & & 10.8cm & {\bf 8.3cm} & & & \\
				\hline  fr1\_desk & 2.5cm & 2.1cm & 3.7cm & 2.0cm & 2.1cm & 2.3cm & 4.3cm & {\bf 1.6cm}\\
				\hline  fr1\_desk2 & 2.8cm & {\bf 2.4cm} & 7.1cm & 4.8cm & 4.6cm & 4.3cm & 4.9cm & \\
				\hline  fr1\_floor & 325.3cm & 26.2cm &  & - & & & & \\
				\hline  fr1\_plant & 5.0cm & 3.8cm & 4.7cm & {\bf 2.2cm} & 2.8cm & 9.1cm & 2.6cm & \\
				\hline  fr1\_room & 14.8cm & 13.4cm & 7.5cm & 6.8cm & {\bf 5.3cm} & 8.4cm & 6.9cm & \\
				\hline  fr1\_rpy & 2.2cm & 3.7cm & 2.8cm & 2.5cm & {\bf 2.0cm} & 2.6cm & 2.7cm & \\
				\hline  fr1\_teddy & 18.7cm & 15.7cm & & 8.3cm & {\bf 3.4cm} & & & \\
				\hline  fr1\_xyz & 1.0cm & {\bf 0.7cm} & 1.7cm & 1.1cm & 1.1cm & 1.4cm & 1.3cm & \\
				\hline  fr2\_360\_hemisphere & 37.6cm & {\bf 31.1cm} & & - & & & & \\								
				\hline  fr2\_360\_kidnap & 132.6cm & {\bf 6.1cm} & & - & & & & \\
				\hline  fr2\_coke & {\bf 17.2cm} & 20.2cm & & - & & & & \\
				\hline  fr2\_desk & 7.2cm & 7.1cm & 3.4cm & 7.1cm & {\bf 1.7cm} & 5.7cm & 5.2cm & \\
				\hline 	fr2\_dishes & 8.4cm & {\bf 7.9cm} & & - & & & & \\
				\hline  fr2\_large\_no\_loop & - & - & & - & & & 8.6cm & \\
				\hline  fr2\_large\_with\_loop & 198.3cm & {\bf 196.7cm} & & - & & & & \\
				\hline  fr2\_metallic\_sphere & {\bf 34.1cm} & 44.3cm & & - & & & & \\
				\hline  fr2\_metallic\_sphere2 & 11.1cm & {\bf 8.4cm} & & - & & & & \\
				\hline  fr2\_pioneer360 & 40.5cm & {\bf 35.8cm} & & - & & & & \\
				\hline  fr2\_pioneer\_slam & 91.2cm & {\bf 85.5cm} & & - & & & & \\
				\hline  fr2\_pioneer\_slam2 & 169.7cm & {\bf 3.3cm} & & - & & & & \\
				\hline  fr2\_pioneer\_slam3 & 28.1cm & {\bf 19.1cm} & & - & & & & \\
				\hline  fr2\_rpy & 0.8cm & {\bf 0.6cm} & & 1.5cm & & & & \\
				\hline  fr2\_xyz & 1.2cm & 1.2cm & 2.9cm & {\bf 1.1cm} & 1.8cm & 0.8cm & 2.0cm & {\bf 1.1cm}\\
				\hline  fr2\_flowerbouquet & 7.0cm & {\bf 5.0cm} & & & & & & \\
				\hline  fr2\_flowerbouquet\_brownbackground & 53.3cm & {\bf 51.7cm} & & & & & & \\
				\hline  fr2\_desk\_with\_person & 4.7cm & {\bf 4.5cm} & & & & & & \\
				\hline  fr3\_cabinet & 39.9cm & {\bf 7.9cm} & & - & & & & \\
				\hline  fr3\_large\_cabinet & 20.9cm & 14.8cm & & {\bf 9.9cm} & & & & \\
				\hline  fr3\_long\_office\_household & 3.2cm & 2.8cm & 3.0cm & {\bf 1.7cm} & 3.5cm & 3.2cm & 4.2cm & 2.2cm\\
				\hline  fr3\_nostructure\_notexture\_far & - & - & & - & & & & \\				
				\hline  fr3\_nostructure\_notexture\_near\_with\_loop & - & - & & - & & & & \\
				\hline  fr3\_nostructure\_texture\_far & 10.8cm & {\bf 5.3cm} & & 7.4cm & & & & \\
				\hline  fr3\_nostructure\_texture\_near\_withloop & 2.9cm & 2.7cm & 3.1cm & 1.6cm & 1.8cm & 1.7cm & 201.8cm & {\bf 1.2cm}\\
				\hline  fr3\_structure\_notexture\_far & - & - & & {\bf 3.0cm} & & & & \\
				\hline  fr3\_structure\_notexture\_near &- & - & & {\bf 2.1cm} & & & & \\
				\hline 	fr3\_structure\_texture\_far & 1.8cm & 1.6cm & & {\bf 1.3cm} & & & & \\
				\hline  fr3\_structure\_texture\_near & 1.6cm & 1.8cm & & {\bf 1.5cm} & & & & \\
				\hline  fr3\_nostructure\_notexture\_near & - & - & & & & & & \\
				\hline  fr3\_teddy & - & - & & {\bf 4.9cm} & & & & \\
				\hline  fr3\_sitting\_xyz & 2.1cm & {\bf 1.7cm} & & & & & & \\
				\hline  fr3\_walking\_xyz & 2.8cm & {\bf 2.4cm} & & & & & & \\
				\hline  fr3\_sitting\_halfsphere & {\bf 1.7cm} & 1.9cm & & & & & & \\
				\hline  fr3\_sitting\_static & 1.3cm & {\bf 0.9cm} & & & & & & \\
				\hline  fr3\_walking\_static & 5.2cm & {\bf 3.9cm} & & & & & & \\
				\hline  fr3\_walking\_rpy & 42.0cm & {\bf 33.7cm} & & & & & & \\
				\hline  fr3\_sitting\_rpy & {\bf 2.7cm} & 4.1cm & & & & & & \\
				\hline  fr3\_walking\_halfsphere & 25.6cm & {\bf 18.2cm} & & & & & & \\
				\hline
			\end{tabular}
			\normalsize
		\end{center}
		\label{tab:error-TUM}}
\end{table*}

\subsection{Qualitative Evaluation}
We first evaluate our system with some challenging datasets captured by ourselves, which may contain complex loops with fast motion and are very long.

{\bf Loop Closure and Low-frame-Rate Sequences.} Figure~\ref{fig:loop-closure} shows an indoor example ``Cubes'' where the scale is large and there are complex loops. The number of frames is 14,817. As can be seen, our method faithfully detect and close the loops, achieving a drift-free 3D reconstruction result, as shown in Figure~\ref{fig:loop-closure}. Our system also can handle low-frame-rate sequences. We extract every third frame from ``Cubes'' sequence to constitute a new sequence. 
Figure~\ref{fig:low-frame-rate}~(a) shows the reconstruction result by our system, which is comparable to the original one. The reconstructions of Kintinuous and ElasticFusion are shown in Figures~\ref{fig:low-frame-rate}~(b) and (c), both of which have serious drift. Please refer to our supplementary video for more examples and comparison results.

{\bf Time Limitation.} Our system can produce drift-free 3D reconstruction without time limitation in a moderate scale scene since the number of keyframes is not always increased in this case. Figure~\ref{fig:long-time} shows another indoor example where the camera capture 20,862 frames in total. Our system can process all data and produce drift-free reconstruction.

{\bf Relocalization.} For some extremely challenging cases, the tracking may be lost. In our system, if the tracking is poor or even lost, the depth map will be not integrated. The camera can be relocalized when the camera moves back to a previously visited position. Please refer to our supplementary video to watch the result and comparison to other systems.

\subsection{Quantitative Evaluation of Trajectory Accuracy}

We use the RGB-D benchmark of Sturm~\etal~\cite{sturm12iros} to evaluate our system and make comparisons with other state-of-the-art systems, i.e.~DVO-SLAM~\cite{KerlSC13}, RGBD-SLAM~\cite{EndresHESCB12}, MRSMap~\cite{StucklerB14}, Kintinuous~\cite{Whelan12rssw}, ElasticFusion~\cite{WhelanSGDL16}, BundleFusion~\cite{dai2017bundle}. We test all scenes in the RGB-D benchmark of Sturm~\etal~\cite{sturm12iros}. Table~\ref{tab:error-TUM} shows the measured absolute trajectory error~(ATE). For other methods, we directly use the reported ATE from their papers: ``-'' indicates tracking failure, and the blank indicates not reported.
Since our system uses keyframes, we compute the RMSE of keyframes for our method. For more fair comparison, we also compute ATE for all frames. Specifically, we output the camera pose when a frame is processed, which will not be further refined by BA. In contrast, the camera poses of keyframes are refined by BA, so their error is further minimized.
As can be seen, our system achieves quite comparable results with the state-of-the-art methods. Compared to ElasticFusion~\cite{WhelanSGDL16} which tested on all of the static scenes in the RGB-D benchmark of Sturm~\etal~\cite{sturm12iros}, our method can track successfully in more scenes, which demonstrate the robustness of our keyframe-based tracking.

\subsection{Quantitative Evaluation of Surface Accuracy}

We perform surface reconstruction accuracy on the synthetic ICL-NUIM dataset~\cite{HandaWMD14}, which provides the synthetic 3D model with ground truth camera poses.
We select three living room scenes (including synthetic noise) to evaluate our approach. The measured ATE RMSE are listed in Table~\ref{tab:error-ICL-NUIM}~\footnote{For ``kt0'' sequence, because 30.5\% frames are lost for our method, we only use the successfully recovered cameras poses to compute ATE RMSE.}, and the surfce reconstruction accuracy results are shown in Table~\ref{tab:error-surface}. For these three scenes, some frames are quite textureless which make tracking very challenging. Especially, our method only perform low-resolution RGB-D alignment and may have problem if the scene is extremely textureless and there are not sufficient features can be matched. Especially, in ``kt0'' sequence, the camera poses of 30.5\% frames are not recovered due to this reason. Due to imperfect camera parameters in this dataset, the surface accuracy of our method is slightly worse than ElasticFusion and BundleFusion.
Nevertheless, the surface accuracy of reconstruction results by using keyframe-based fusion and every frame fusion are quite comparable, which demonstrate the effectiveness of our keyframe-based fusion method.

\begin{table} [tb!]
	{
		\caption{Comparison of ATE RMSE on the synthetic ICL-NUIM dataset.}
		\begin{center}
			\footnotesize
			\begin{tabular}{|c|c|c|c|}
	\hline   Seq. & kt0 & kt1 & kt2 \\
	\hline  Ours~(all frames) & 11.9cm & 2.4cm & 4.2cm \\
	\hline  Ours~(key frames) & 1.8cm & 1.6cm & 3.2cm \\
	\hline  Kintinuous & 7.2cm & 0.5cm & 1.0cm \\
	\hline  ElasticFusion & 0.9cm & 0.9cm & 1.4cm\\				
	\hline  DVO-SLAM & 10.4cm & 2.9cm & 19.1cm\\				
	\hline  RGB-D SLAM & 2.6cm & 0.8cm & 1.8cm\\			
	\hline  MRSMap & 20.4cm & 22.8cm & 18.9cm\\			
	\hline  BundleFusion & {\bf 0.6cm} & {\bf 0.4cm} & {\bf 0.6cm}\\												
	\hline
\end{tabular}
			\normalsize
		\end{center}
		\label{tab:error-ICL-NUIM}}
\end{table}

\begin{table} [tb!]
	{
		\caption{Comparison of surface accuracy on the synthetic ICL-NUIM dataset.}
		\begin{center}
			\footnotesize
			\begin{tabular}{|c|c|c|c|}
				\hline   Seq. & kt0 & kt1 & kt2 \\
				\hline  Our keyframe-based fusion  & 0.9cm & 1.1cm & 1.6cm \\
				\hline  Our every frame fusion & 0.8cm & 1.3cm & 1.8cm\\
				\hline  Kintinuous & 1.1cm & 0.8cm & 0.9cm\\
				\hline  ElasticFusion & 0.7cm & 0.7cm & 0.8cm\\				
				\hline  DVO-SLAM & 3.2cm & 6.1cm & 11.9cm\\				
				\hline  RGB-D SLAM & 4.4cm & 3.2cm & 3.1cm\\				
				\hline  MRSMap & 6.1cm & 14.0cm & 9.8cm\\							
				\hline  BundleFusion & {\bf 0.5cm} & {\bf 0.6cm} & {\bf 0.7cm}\\												
				\hline
			\end{tabular}
			\normalsize
		\end{center}
		\label{tab:error-surface}}
\end{table}

\section{Discussions and Conclusions}
In this paper, we have presented a novel keyframe-based dense SLAM approach which is not only robust to fast motion, but also can recover from tracking failure, handle loop closure and adjust the dense surface online to achieve drift-free 3D reconstruction even on a laptop. In order to achieve this goal, we first contributed a keyframe-based tracking approach which combines color and geometry information to make the tracking as robust as possible. Secondly, we proposed a novel incremental BA which makes maximal use of intermediate computation to save computation and adaptively update necessary keyframes for map refinement. Finally, we proposed a keyframe-based dense mapping method which can adjust the dense surface online by a few de-integration and integration operations. The depths of non-keyframes are fused into keyframes as much as possible to reduce redundancy.
With this representation, our system not only can adjust the dense surface online but also can operate for extended periods of time in a moderate size scene.

Our system still has some limitations. If the scene is fully planar and extremely textureless, our system may fail. In addition, our system still has difficulties in handling serious motion blur. In future work, we would like to include inertial measurements to further increase robustness.


%

\section*{Acknowledgment}
The authors would like to thank Bangbang Yang, Weijian Xie and Shangjin Zhai for their kind help in making experimental results and comparisons.

\ifCLASSOPTIONcaptionsoff
  \newpage
\fi



\bibliographystyle{IEEEtran}
\bibliography{reference}

\begin{thebibliography}{10}
\providecommand{\url}[1]{#1}
\csname url@samestyle\endcsname
\providecommand{\newblock}{\relax}
\providecommand{\bibinfo}[2]{#2}
\providecommand{\BIBentrySTDinterwordspacing}{\spaceskip=0pt\relax}
\providecommand{\BIBentryALTinterwordstretchfactor}{4}
\providecommand{\BIBentryALTinterwordspacing}{\spaceskip=\fontdimen2\font plus
\BIBentryALTinterwordstretchfactor\fontdimen3\font minus
  \fontdimen4\font\relax}
\providecommand{\BIBforeignlanguage}[2]{{%
\expandafter\ifx\csname l@#1\endcsname\relax
\typeout{** WARNING: IEEEtran.bst: No hyphenation pattern has been}%
\typeout{** loaded for the language `#1'. Using the pattern for}%
\typeout{** the default language instead.}%
\else
\language=\csname l@#1\endcsname
\fi
#2}}
\providecommand{\BIBdecl}{\relax}
\BIBdecl

\bibitem{klein2007parallel}
G.~Klein and D.~W. Murray, ``Parallel tracking and mapping for small {AR}
  workspaces,'' in \emph{6th {IEEE/ACM} International Symposium on Mixed and
  Augmented Reality}, 2007, pp. 225--234.

\bibitem{engel2014lsd}
J.~Engel, T.~Sch{\"o}ps, and D.~Cremers, ``{LSD-SLAM}: Large-scale direct
  monocular {SLAM},'' in \emph{13th European Conference on Computer Vision,
  Part {II}}.\hskip 1em plus 0.5em minus 0.4em\relax Springer, 2014, pp.
  834--849.

\bibitem{mur2015orb}
R.~Mur-Artal, J.~Montiel, and J.~D. Tardos, ``{ORB-SLAM}: a versatile and
  accurate monocular {SLAM} system,'' \emph{IEEE Transactions on Robotics},
  vol.~31, no.~5, pp. 1147--1163, 2015.

\bibitem{NewcombeIHMKDKSHF11}
R.~A. Newcombe, S.~Izadi, O.~Hilliges, D.~Molyneaux, D.~Kim, A.~J. Davison,
  P.~Kohli, J.~Shotton, S.~Hodges, and A.~W. Fitzgibbon, ``Kinectfusion:
  Real-time dense surface mapping and tracking,'' in \emph{10th {IEEE}
  International Symposium on Mixed and Augmented Reality}, 2011, pp. 127--136.

\bibitem{NiessnerZIS13}
M.~Nie{\ss}ner, M.~Zollh{\"{o}}fer, S.~Izadi, and M.~Stamminger, ``Real-time
  {3D} reconstruction at scale using voxel hashing,'' \emph{{ACM} Trans.
  Graph.}, vol.~32, no.~6, pp. 169:1--169:11, 2013.

\bibitem{WhelanLSGD15}
T.~Whelan, S.~Leutenegger, R.~F. Salas{-}Moreno, B.~Glocker, and A.~J. Davison,
  ``Elasticfusion: Dense {SLAM} without {A} pose graph,'' in \emph{Robotics:
  Science and Systems XI, Sapienza University of Rome, Rome, Italy, July 13-17,
  2015}, 2015.

\bibitem{WhelanSGDL16}
T.~Whelan, R.~F. Salas{-}Moreno, B.~Glocker, A.~J. Davison, and S.~Leutenegger,
  ``Elasticfusion: Real-time dense {SLAM} and light source estimation,''
  \emph{I. J. Robotics Res.}, vol.~35, no.~14, pp. 1697--1716, 2016.

\bibitem{InfiniTAM_ISMAR_2015}
O.~Kahler, V.~A. Prisacariu, C.~Y. Ren, X.~Sun, P.~H.~S. Torr, and D.~W.
  Murray, ``{Very High Frame Rate Volumetric Integration of Depth Images on
  Mobile Device},'' \emph{{IEEE Transactions on Visualization and Computer
  Graphics (Proceedings of ISMAR 2015)}}, vol.~22, no.~11, 2015.

\bibitem{WhelanKJFLM15}
T.~Whelan, M.~Kaess, H.~Johannsson, M.~F. Fallon, J.~J. Leonard, and
  J.~McDonald, ``Real-time large-scale dense {RGB-D} {SLAM} with volumetric
  fusion,'' \emph{I. J. Robotics Res.}, vol.~34, no. 4-5, pp. 598--626, 2015.

\bibitem{dai2017bundle}
A.~Dai, M.~Nie{\ss}ner, M.~Zollh{\"{o}}fer, S.~Izadi, and C.~Theobalt,
  ``Bundlefusion: Real-time globally consistent {3D} reconstruction using
  on-the-fly surface reintegration,'' \emph{{ACM} Trans. Graph.}, vol.~36,
  no.~3, pp. 24:1--24:18, 2017.

\bibitem{MaierSC17}
R.~Maier, R.~Schaller, and D.~Cremers, ``Efficient online surface correction
  for real-time large-scale {3D} reconstruction,'' in \emph{BMVC}, 2017, pp.
  1--12.

\bibitem{KerlSC13}
C.~Kerl, J.~Sturm, and D.~Cremers, ``Dense visual {SLAM} for {RGB-D} cameras,''
  in \emph{{IEEE/RSJ} International Conference on Intelligent Robots and
  Systems}, 2013, pp. 2100--2106.

\bibitem{HuangBHKMFR11}
A.~S. Huang, A.~Bachrach, P.~Henry, M.~Krainin, D.~Maturana, D.~Fox, and
  N.~Roy, ``Visual odometry and mapping for autonomous flight using an {RGB-D}
  camera,'' in \emph{The 15th International Symposium on Robotics Research},
  2011, pp. 235--252.

\bibitem{endres20143}
F.~Endres, J.~Hess, J.~Sturm, D.~Cremers, and W.~Burgard, ``{3-D} mapping with
  an {RGB-D} camera,'' \emph{IEEE Transactions on Robotics}, vol.~30, no.~1,
  pp. 177--187, 2014.

\bibitem{BeslM92}
P.~J. Besl and N.~D. McKay, ``A method for registration of {3-D} shapes,''
  \emph{{IEEE} Trans. Pattern Anal. Mach. Intell.}, vol.~14, no.~2, pp.
  239--256, 1992.

\bibitem{Whelan12rssw}
T.~Whelan, J.~McDonald, M.~Kaess, M.~Fallon, H.~Johannsson, and J.~Leonard,
  ``Kintinuous: Spatially extended {K}inect{F}usion,'' in \emph{RSS Workshop on
  RGB-D: Advanced Reasoning with Depth Cameras}, Sydney, Australia, Jul 2012.

\bibitem{ZengZZL13}
M.~Zeng, F.~Zhao, J.~Zheng, and X.~Liu, ``Octree-based fusion for realtime {3D}
  reconstruction,'' \emph{Graphical Models}, vol.~75, no.~3, pp. 126--136,
  2013.

\bibitem{LiuZB16}
H.~Liu, G.~Zhang, and H.~Bao, ``Robust keyframe-based monocular {SLAM} for
  augmented reality,'' in \emph{{IEEE} International Symposium on Mixed and
  Augmented Reality}, 2016.

\bibitem{NewcombeLD11}
R.~A. Newcombe, S.~Lovegrove, and A.~J. Davison, ``{DTAM:} dense tracking and
  mapping in real-time,'' in \emph{{IEEE} International Conference on Computer
  Vision}, 2011, pp. 2320--2327.

\bibitem{PradeepRIZBB13}
V.~Pradeep, C.~Rhemann, S.~Izadi, C.~Zach, M.~Bleyer, and S.~Bathiche,
  ``Monofusion: Real-time {3D} reconstruction of small scenes with a single web
  camera,'' in \emph{{IEEE} International Symposium on Mixed and Augmented
  Reality}, 2013, pp. 83--88.

\bibitem{SchopsSHP15}
T.~Sch{\"{o}}ps, T.~Sattler, C.~H{\"{a}}ne, and M.~Pollefeys, ``{3D} modeling
  on the go: Interactive {3D} reconstruction of large-scale scenes on mobile
  devices,'' in \emph{2015 International Conference on 3D Vision}, 2015, pp.
  291--299.

\bibitem{OndruskaKI15}
P.~Ondruska, P.~Kohli, and S.~Izadi, ``Mobilefusion: Real-time volumetric
  surface reconstruction and dense tracking on mobile phones,'' \emph{{IEEE}
  Trans. Vis. Comput. Graph.}, vol.~21, no.~11, pp. 1251--1258, 2015.

\bibitem{ZhouMK13}
Q.~Zhou, S.~Miller, and V.~Koltun, ``Elastic fragments for dense scene
  reconstruction,'' in \emph{{IEEE} International Conference on Computer
  Vision, {ICCV} 2013, Sydney, Australia, December 1-8, 2013}, 2013, pp.
  473--480.

\bibitem{WangZB14}
K.~Wang, G.~Zhang, and H.~Bao, ``Robust {3D} reconstruction with an {RGB-D}
  camera,'' \emph{{IEEE} Trans. Image Processing}, vol.~23, no.~11, pp.
  4893--4906, 2014.

\bibitem{ChoiZK15}
S.~Choi, Q.~Zhou, and V.~Koltun, ``Robust reconstruction of indoor scenes,'' in
  \emph{{IEEE} Conference on Computer Vision and Pattern Recognition}, 2015,
  pp. 5556--5565.

\bibitem{Triggs99}
B.~Triggs, P.~F. McLauchlan, R.~I. Hartley, and A.~W. Fitzgibbon, ``Bundle
  adjustment - a modern synthesis.'' in \emph{Workshop on Vision Algorithms},
  1999, pp. 298--372.

\bibitem{OlsonLT06}
E.~Olson, J.~J. Leonard, and S.~J. Teller, ``Fast iterative alignment of pose
  graphs with poor initial estimates,'' in \emph{Proceedings of {IEEE}
  International Conference on Robotics and Automation}, 2006, pp. 2262--2269.

\bibitem{StrasdatMD10}
H.~Strasdat, J.~M.~M. Montiel, and A.~J. Davison, ``Scale drift-aware large
  scale monocular {SLAM},'' in \emph{Robotics: Science and Systems VI}, 2010.

\bibitem{KummerleGSKB11}
R.~K{\"{u}}mmerle, G.~Grisetti, H.~Strasdat, K.~Konolige, and W.~Burgard,
  ``g\({}^{\mbox{2}}\)o: {A} general framework for graph optimization,'' in
  \emph{{IEEE} International Conference on Robotics and Automation}, 2011, pp.
  3607--3613.

\bibitem{kaess2008isam}
M.~Kaess, A.~Ranganathan, and F.~Dellaert, ``{iSAM}: Incremental smoothing and
  mapping,'' \emph{IEEE Transactions on Robotics}, vol.~24, no.~6, pp.
  1365--1378, 2008.

\bibitem{kaess2012isam2}
M.~Kaess, H.~Johannsson, R.~Roberts, V.~Ila, J.~J. Leonard, and F.~Dellaert,
  ``{iSAM2}: Incremental smoothing and mapping using the bayes tree,''
  \emph{The International Journal of Robotics Research}, vol.~31, no.~2, pp.
  216--235, 2012.

\bibitem{ila2017slam++}
V.~Ila, L.~Polok, M.~Solony, and P.~Svoboda, ``{SLAM++} -a highly efficient and
  temporally scalable incremental slam framework,'' \emph{The International
  Journal of Robotics Research}, vol.~36, no.~2, pp. 210--230, 2017.

\bibitem{IlaPSI17}
V.~Ila, L.~Polok, M.~Solony, and K.~Istenic, ``Fast incremental bundle
  adjustment with covariance recovery,'' in \emph{International Conference on
  3D Vision}, 2017, pp. 4321--4330.

\bibitem{rublee2011orb}
E.~Rublee, V.~Rabaud, K.~Konolige, and G.~Bradski, ``{ORB}: an efficient
  alternative to {SIFT} or {SURF},'' in \emph{IEEE International Conference on
  Computer Vision}.\hskip 1em plus 0.5em minus 0.4em\relax IEEE, 2011, pp.
  2564--2571.

\bibitem{Galvez-LopezT12}
D.~G{\'{a}}lvez{-}L{\'{o}}pez and J.~D. Tard{\'{o}}s, ``Bags of binary words
  for fast place recognition in image sequences,'' \emph{{IEEE} Transactions on
  Robotics}, vol.~28, no.~5, pp. 1188--1197, 2012.

\bibitem{klein2008improving}
G.~Klein and D.~W. Murray, ``Improving the agility of keyframe-based {SLAM},''
  in \emph{10th European Conference on Computer Vision, Part {II}}.\hskip 1em
  plus 0.5em minus 0.4em\relax Springer, 2008, pp. 802--815.

\bibitem{VogiatzisH11}
G.~Vogiatzis and C.~Hern{\'{a}}ndez, ``Video-based, real-time multi-view
  stereo,'' \emph{Image Vision Comput.}, vol.~29, no.~7, pp. 434--441, 2011.

\bibitem{forster2014svo}
C.~Forster, M.~Pizzoli, and D.~Scaramuzza, ``{SVO}: Fast semi-direct monocular
  visual odometry,'' in \emph{IEEE International Conference on Robotics and
  Automation}.\hskip 1em plus 0.5em minus 0.4em\relax IEEE, 2014, pp. 15--22.

\bibitem{PizzoliFS14}
M.~Pizzoli, C.~Forster, and D.~Scaramuzza, ``{REMODE:} probabilistic, monocular
  dense reconstruction in real time,'' in \emph{2014 {IEEE} International
  Conference on Robotics and Automation, {ICRA} 2014, Hong Kong, China, May 31
  - June 7, 2014}, 2014, pp. 2609--2616.

\bibitem{civera2008inverse}
J.~Civera, A.~J. Davison, and J.~M. Montiel, ``Inverse depth parametrization
  for monocular {SLAM},'' \emph{IEEE Transactions on Robotics}, vol.~24, no.~5,
  pp. 932--945, 2008.

\bibitem{sturm12iros}
J.~Sturm, N.~Engelhard, F.~Endres, W.~Burgard, and D.~Cremers, ``A benchmark
  for the evaluation of {RGB-D} {SLAM} systems,'' in \emph{{IEEE/RSJ}
  International Conference on Intelligent Robot Systems}, Oct. 2012, pp.
  573--580.

\bibitem{hartley2003multiple}
R.~Hartley and A.~Zisserman, \emph{Multiple view geometry in computer
  vision}.\hskip 1em plus 0.5em minus 0.4em\relax Cambridge university press,
  2004.

\bibitem{jeong2012pushing}
Y.~Jeong, D.~Nister, D.~Steedly, R.~Szeliski, and I.-S. Kweon, ``Pushing the
  envelope of modern methods for bundle adjustment,'' \emph{IEEE transactions
  on pattern analysis and machine intelligence}, vol.~34, no.~8, pp.
  1605--1617, 2012.

\bibitem{kaess2011isam2}
M.~Kaess, H.~Johannsson, R.~Roberts, V.~Ila, J.~J. Leonard, and F.~Dellaert,
  ``{iSAM2}: Incremental smoothing and mapping using the bayes tree,''
  \emph{International Journal of Robotics Research}, vol.~31, no.~2, pp.
  216--235, 2012.

\bibitem{EndresHESCB12}
F.~Endres, J.~Hess, N.~Engelhard, J.~Sturm, D.~Cremers, and W.~Burgard, ``An
  evaluation of the {RGB-D} {SLAM} system,'' in \emph{{IEEE} International
  Conference on Robotics and Automation, {ICRA} 2012, 14-18 May, 2012, St.
  Paul, Minnesota, {USA}}, 2012, pp. 1691--1696.

\bibitem{StucklerB14}
J.~St{\"{u}}ckler and S.~Behnke, ``Multi-resolution surfel maps for efficient
  dense {3D} modeling and tracking,'' \emph{J. Visual Communication and Image
  Representation}, vol.~25, no.~1, pp. 137--147, 2014.

\bibitem{HandaWMD14}
A.~Handa, T.~Whelan, J.~McDonald, and A.~J. Davison, ``A benchmark for {RGB-D}
  visual odometry, {3D} reconstruction and {SLAM},'' in \emph{2014 {IEEE}
  International Conference on Robotics and Automation, {ICRA} 2014, Hong Kong,
  China, May 31 - June 7, 2014}, 2014, pp. 1524--1531.

\end{thebibliography}
\end{document}